\documentclass[11pt]{article}

\usepackage[]{acl}

\usepackage{times}
\usepackage{latexsym}

\usepackage[T1]{fontenc}

\usepackage[utf8]{inputenc}

\usepackage{microtype}

\usepackage{inconsolata}

\usepackage{graphicx}
\usepackage{hyperref}
\usepackage{url}
\usepackage{amsmath}
\usepackage{amsfonts}
\usepackage{graphicx}
\usepackage{subcaption}
\usepackage{bm}
\usepackage{booktabs}
\usepackage{caption}
\usepackage{multirow}
\usepackage{array}
\usepackage{enumitem}
\usepackage{algorithm}
\usepackage{algpseudocode}
\usepackage{cleveref}
\usepackage{stfloats}
\usepackage{xcolor}
\usepackage{tcolorbox}
\usepackage{colortbl}
\usepackage{textcomp}
\usepackage{tabularx}
\usepackage{dashrule}
\usepackage{relsize}

\newcommand{\think}[1]{\textcolor{blue!50!white}{\textbf{\texttt{<#1>}}}}
\newcommand{\passage}[0]{\textcolor{orange!50!white}{\textbf{\texttt{[passage]}}}}
\newcommand{\graph}[0]{\textcolor{red!50!white}{\textbf{\texttt{[graph]}}}}
\newcommand{\search}[1]{\textcolor{violet!50!white}{\textbf{\texttt{<#1>}}}}
\newcommand{\info}[1]{\textcolor{brown!50!white}{\textbf{\texttt{<#1>}}}}
\newcommand{\answer}[1]{\textcolor{teal!50!white}{\textbf{\texttt{<#1>}}}}

\definecolor{lavender}{RGB}{230,230,250}
\definecolor{brickred}{RGB}{150,0,24}

\algrenewcommand{\algorithmiccomment}[1]{\hfill\textcolor{blue!50!white}{\small $\triangleright$ #1}}
\algtext*{EndIf}
\algtext*{EndWhile}

\title{{\centering\raisebox{-.3\height}{\includegraphics[width=0.7cm]{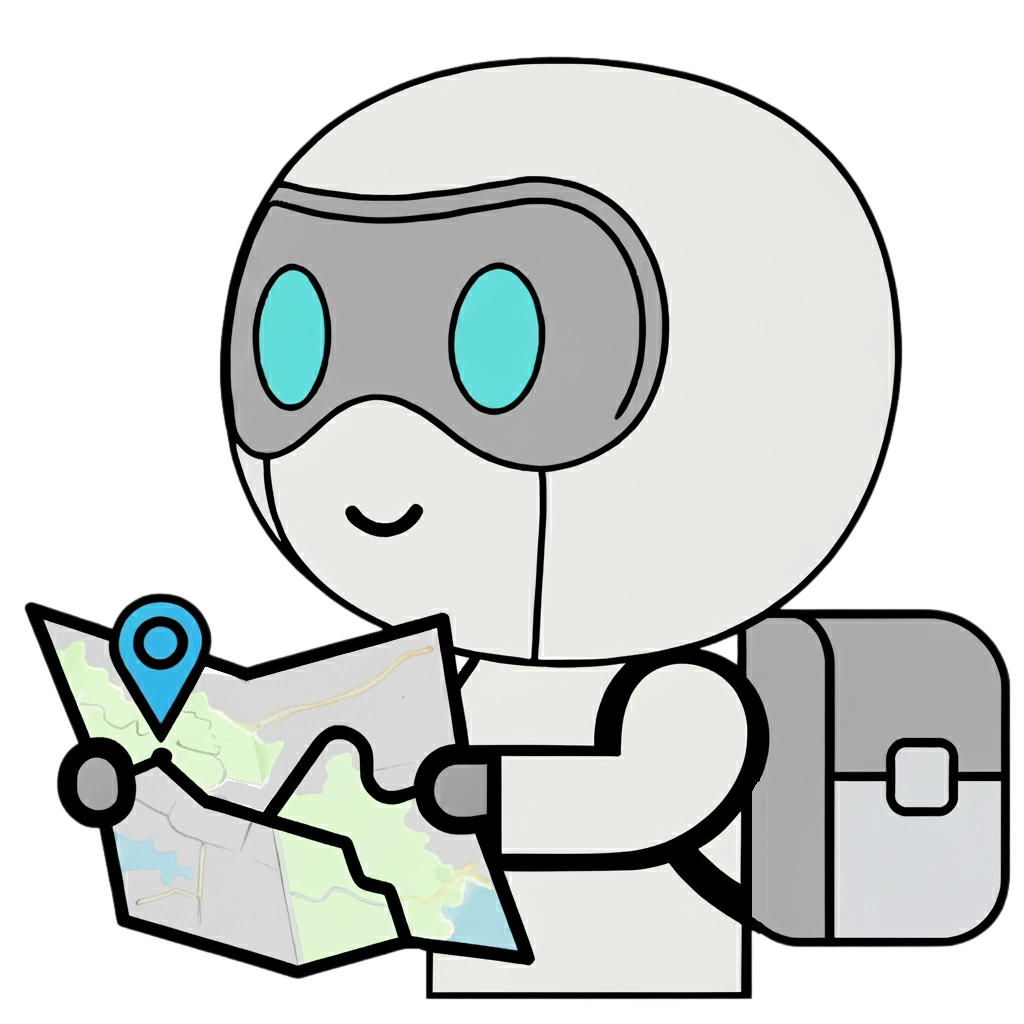}}} RouteRAG: Efficient Retrieval-Augmented Generation from Text and Graph via Reinforcement Learning}

\author{
    Yucan Guo,
    Miao Su,
    \setcounter{footnote}{1}
    Saiping Guan\thanks{Corresponding authors.},
    Zihao Sun,
    Xiaolong Jin$^{\dagger}$,
    Jiafeng Guo,
    Xueqi Cheng
     \\
    \textsuperscript{1}CAS Key Laboratory of Network Data Science and Technology, \\ Institute of Computing Technology, Chinese Academy of Sciences \\
    \textsuperscript{2}School of Computer Science and Technology, University of Chinese Academy of Sciences \\
    \small
    \texttt{\{guoyucan23z, sumiao22z, guansaiping, sunzihao18z, jinxiaolong, guojiafeng, cxq\}@ict.ac.cn}\\
}

\newcommand{\model}{RouteRAG}

\begin{document}
\maketitle
\begin{abstract}
Retrieval-Augmented Generation (RAG) integrates non-parametric knowledge into Large Language Models (LLMs), typically from unstructured texts and structured graphs. While recent progress has advanced text-based RAG to multi-turn reasoning through Reinforcement Learning (RL), extending these advances to hybrid retrieval introduces additional challenges. 
Existing graph-based or hybrid systems typically depend on fixed or handcrafted retrieval pipelines, lacking the ability to integrate supplementary evidence as reasoning unfolds. Besides, while graph evidence provides relational structures crucial for multi-hop reasoning, it is substantially more expensive to retrieve. 
To address these limitations, we introduce \model{}, an RL-based framework that enables LLMs to perform multi-turn and adaptive graph-text hybrid RAG. \model{} jointly optimizes the entire generation process via RL, allowing the model to learn when to reason, what to retrieve from either texts or graphs, and when to produce final answers, all within a unified generation policy.
To guide this learning process, we design a two-stage training framework that accounts for both task outcome and retrieval efficiency, enabling the model to exploit hybrid evidence while avoiding unnecessary retrieval overhead.
Experimental results across five question answering benchmarks demonstrate that \model{} significantly outperforms existing RAG baselines, highlighting the benefits of end-to-end RL in supporting adaptive and efficient retrieval for complex reasoning. \footnote{The code is publicly available at \url{https://github.com/YucanGuo/RouteRAG}.} 
\end{abstract}

\section{Introduction}
Large Language Models (LLMs) have demonstrated remarkable capabilities in reasoning, decision-making, and long-form generation~\citep{Zhao2023survey, Touvron2023llama, Team2024gemini}, especially when further trained with Reinforcement Learning (RL)~\citep{Achiam2023gpt, Guo2025deepseek, Yang2025qwen3}. 
These abilities have enabled LLMs to follow complex instructions, emulate chain-of-thought reasoning, and solve complicated multi-hop questions~\citep{Zhou2023instruction, Wei2022chain}. 
However, the knowledge of LLMs remains static, bounded by the data available at pretraining time. As a result, LLMs often produce inaccurate or outdated outputs when faced with knowledge-intensive queries that require access to external or up-to-date information~\citep{Augenstein2024factuality, Huang2025survey}.

To overcome this limitation, Retrieval-Augmented Generation (RAG) has emerged as a core paradigm for enhancing LLMs with access to external knowledge sources~\citep{Lewis2020retrieval, Gao2023retrieval}. 
Early RAG systems typically perform a single round of retrieval before generation~\citep{Guu2020retrieval, Wang2023self}. Recent work has shown the benefits of multi-turn retrieval, where the model interleaves retrieval and reasoning over multiple steps~\citep{Yao2023react, Trivedi2023interleaving, Li2025search}. 
However, these prompt-based approaches often depend on large closed-source models with strong intrinsic reasoning and planning skills. Smaller open-source models struggle to determine when to retrieve, how to formulate retrieval queries, and how to analyze retrieved evidence. 
This gap has motivated a new line of research~\citep{Jin2025search, Song2025r1} that employs RL to explicitly train models to make retrieval and reasoning decisions. By optimizing a learned policy over interleaved thinking and retrieval actions, these RL-based methods aim to equip models with adaptive, context-sensitive retrieval strategies that surpass static instructions.

In parallel, graph-based RAG systems~\citep{Edge2024local, Jimenez2024hipporag, Gutierrez2025from} utilize structured knowledge graphs to integrate and reason over information scattered across multiple passages, thereby improving coverage of factual entities and relations. 
While graphs enable more accurate entity disambiguation and multi-hop path reasoning than text-only retrieval, retrieving and processing graph evidence is often more computationally expensive, especially in large-scale or dense graphs.
Moreover, existing graph-based RAG systems typically operate in a one-shot retrieval setting, fetching graph evidence once before generation, and lack the ability to adaptively choose between graph and text retrieval based on the evolving information needs of the query. 
Consequently, the current architecture of graph-based RAG systems presents challenges in managing complex reasoning that necessitates multi-turn interactions, and can also lead to unnecessary retrieval overhead when reasoning chain is too long.

We address these limitations with \model{}, an RL-based framework that enables LLMs to perform multi-turn and hybrid retrieval over both unstructured texts and structured knowledge graphs. Instead of passively executing preset instructions, \model{} actively orchestrates retrieval decisions, selecting when and where to access external knowledge.
To overcome the challenges of managing complex reasoning and avoiding unnecessary retrieval overhead, \model{} learns to interleave reasoning, retrieval, and answer formulation through a unified generation policy, adapting its retrieval behavior to the evolving task context.

To enable \model{} to generate accurate answers while efficiently retrieving relevant knowledge, we adopt a two-stage Group Relative Policy Optimization (GRPO)~\citep{Shao2024deepseekmath} training framework. In the first stage, the model is rewarded solely for answer correctness, allowing it to acquire the core capability of generating accurate responses and establishing a solid starting point for further optimization. In the second stage, we introduce an additional efficiency reward that discourages unnecessary retrieval, guiding the model to strike a balance between accuracy and computational cost. 
With these designs, \model{} can achieve both high accuracy and retrieval efficiency in complex multi-hop reasoning tasks.

Our main contributions lie in three aspects:
\begin{itemize}
    \item We propose \model{}, an RL-based framework for multi-turn and hybrid RAG. The model learns a unified generation policy that interleaves reasoning, adaptive graph-text hybrid retrieval, and answer formulation through a two-stage training framework.
    \item We design a reward function that jointly optimizes answer accuracy and retrieval efficiency, encouraging the model to retrieve selectively and to reason effectively over retrieved evidence across multiple steps.
    \item Extensive experiments on five Question Answering (QA) benchmarks demonstrate that \model{} outperforms prior multi-turn and graph-based RAG systems significantly.
\end{itemize}

\section{Related Work}
\subsection{RAG}
RAG has become a key paradigm for enhancing LLMs with external knowledge, thus mitigating hallucination and improving factual grounding~\citep{Guu2020retrieval, Gao2023retrieval}. Traditional RAG systems retrieve relevant text chunks from an external knowledge base according to the query, and then feed the query into the LLM together with those text chunks to generate a final answer~\citep{Lewis2020retrieval, Yu2022survey}. 
Beyond such one-shot retrieve-then-generate pipelines, recent research has explored multi-turn retrieval to provide more fine-grained and incremental supplementation of external knowledge, interleaving reasoning with evidence acquisition. For instance, IRCoT~\citep{Trivedi2023interleaving} shows that alternating chain-of-thought reasoning with retrieval improves the performance of LLM on knowledge-intensive multi-hop QA. Search-o1~\citep{Li2025search} further develops this line by introducing a reason-in-documents module to alleviate the issue of redundant information in retrieved documents.

As an alternative route for deep knowledge integration, graph-based RAG methods incorporate structured knowledge graphs to aggregate evidence across passages and to make relational connections explicit~\citep{Peng2024graph, Zhao2023survey}. By exposing entities and relations directly, these methods are particularly effective for multi-hop questions that require linking facts across disparate documents~\citep{Edge2024local, Jimenez2024hipporag, Gutierrez2025from}. However, graph retrieval is often more computationally expensive than text retrieval, and existing methods commonly perform one-shot retrieval. 
Although recent work such as HybGRAG~\citep{Lee2025hybgrag} demonstrates that multi-turn hybrid text-graph retrieval is feasible through predefined multi-step procedures, these methods rely on fixed heuristics rather than a learnable policy.

\subsection{RL for LLM Reasoning}
RL has played a central role in improving the reasoning capabilities of LLMs. RL from Human Feedback (RLHF)~\citep{Christiano2017deep, Ouyang2022training} has established a standard paradigm, where a reward model trained from human preferences directs the optimization of policies~\citep{Lambert2025rewardbench}, allowing models to adhere to instructions and reason with greater accuracy. Proximal Policy Optimization (PPO)~\citep{Schulman2017proximal} remains the predominant algorithm for achieving these goals.
More recently, GRPO~\citep{Shao2024deepseekmath} has been proposed as a more efficient variant, which leverages group-wise relative rewards to stabilize training and reduce variance. 
Building on these advances, researchers have begun to apply RL directly to the training of multi-turn RAG systems~\citep{Jin2025search, Song2025r1}. For instance, Search-R1~\citep{Jin2025search} trains LLMs with RL to decide when and what to search in the middle of reasoning, using only outcome rewards. While this reward design effectively improves correctness, it does not explicitly address retrieval cost or efficiency.

\section{\model{}}
In this section, we present \model{}, an RL-based framework for multi-turn hybrid RAG. We first describe the multi-turn workflow and the mechanism for hybrid knowledge access (\Cref{sec:overall_framework}). Subsequently, we introduce our two-stage RL framework (\Cref{sec:two_stage_RL}), encompassing the formulation of outcome and efficiency rewards, alongside the GRPO-based training algorithm.

\begin{algorithm}[htbp]
\small
\caption{\model{} Framework}
\label{alg:model_framework}
\begin{algorithmic}[1]
\Require Input query $q$, policy model $\pi_\theta$, retriever $\mathcal{R}$, maximum step budget $B$.
\Ensure Final response $y$.
\State Initialize response $y \gets \emptyset$, step count $b \gets 0$
\While{$b < B$}
    \State Initialize current rollout $y_b \gets \emptyset$
    \While{True}
        \State Sample next token $y' \sim \pi_\theta(\cdot \mid q, y+y_b)$
        \State $y_b \gets y_b + y'$
        \If{$y' \in$ \{\search{/search}, \answer{/answer}, $<$eos$>$\}}
            \State \textbf{break}
        \EndIf
    \EndWhile
    \State $y \gets y + y_b$ \Comment{Combine rollout with history}
    \If{\search{search}...\search{/search} detected in $y_b$}
        \State \textbf{if} \passage{} in $y_b$ \textbf{then} $m \gets \texttt{Passage}$
        \State \textbf{if} \graph{} in $y_b$ \textbf{then} $m \gets$ \texttt{Graph}
        \If{\passage{} and \graph{} in $y_b$}
            \State $m \gets$ \texttt{Hybrid}
        \EndIf
        \State Extract query $q' \gets \text{ParseQuery}(y_b)$
        \State $d \gets \mathcal{R}(q', m)$ \Comment{Retrieve documents according to the retrieval mode $m$}
        \State $y \gets y + \info{information} d \info{/information}$
    \ElsIf{\answer{answer}...\answer{/answer} detected in $y_b$}
        \State \Return final response $y$
    \EndIf
    \State $b \gets b + 1$
\EndWhile
\State \Return final response $y$
\end{algorithmic}
\end{algorithm}

\subsection{Overall Framework}
\label{sec:overall_framework}
We begin by outlining the overall architecture of \model{}. This framework integrates LLMs with external retrievers in a multi-turn reasoning loop, where special tokens from the reasoning process can trigger retrieval actions from text and graph knowledge sources. We describe the multi-turn reasoning and retrieval workflow in~\Cref{sec:multi_turn_workflow} and the hybrid knowledge access mechanisms of the external retriever in~\Cref{sec:retriever}.

\subsubsection{Multi-Turn Reasoning and Hybrid Retrieval Workflow}
\label{sec:multi_turn_workflow}
We formulate multi-turn retrieval-augmented generation as a sequential decision-making process, as shown in \Cref{alg:model_framework}. Given an input query $q$, the policy model $\pi_\theta$ interacts with external knowledge sources over a sequence of steps $b = \{1, \dots, B\}$, where $B$ is the maximum step budget. At each step, the policy model conditions on the query and the current context to generate an action token. The action space includes continuing internal reasoning, triggering a retrieval operation (\search{search}...\search{/search}), or producing a final answer (\answer{answer}...\answer{/answer}). The retrieval operation further specifies a retrieval mode $m \in \{\texttt{Passage}, \texttt{Graph}, \texttt{Hybrid}\}$ by special tokens $y \in \{\passage{},\graph{}\}$ and a sub-query $q'$, which are used to obtain documents $d$ from the retriever $\mathcal{R}(q', m)$. The retrieved information is then appended to the context and becomes available for subsequent reasoning.

This workflow enables the model to progressively refine its knowledge state by deciding what to retrieve and when to retrieve it, conditioned on the evolving reasoning trajectory. Moreover, the explicit action space over different retrieval modes allows the model to adaptively choose among passage, graph, and hybrid retrieval, depending on the requirements of the query.

\begin{figure*}[tbp]
    \centering
    \includegraphics[width=\linewidth]{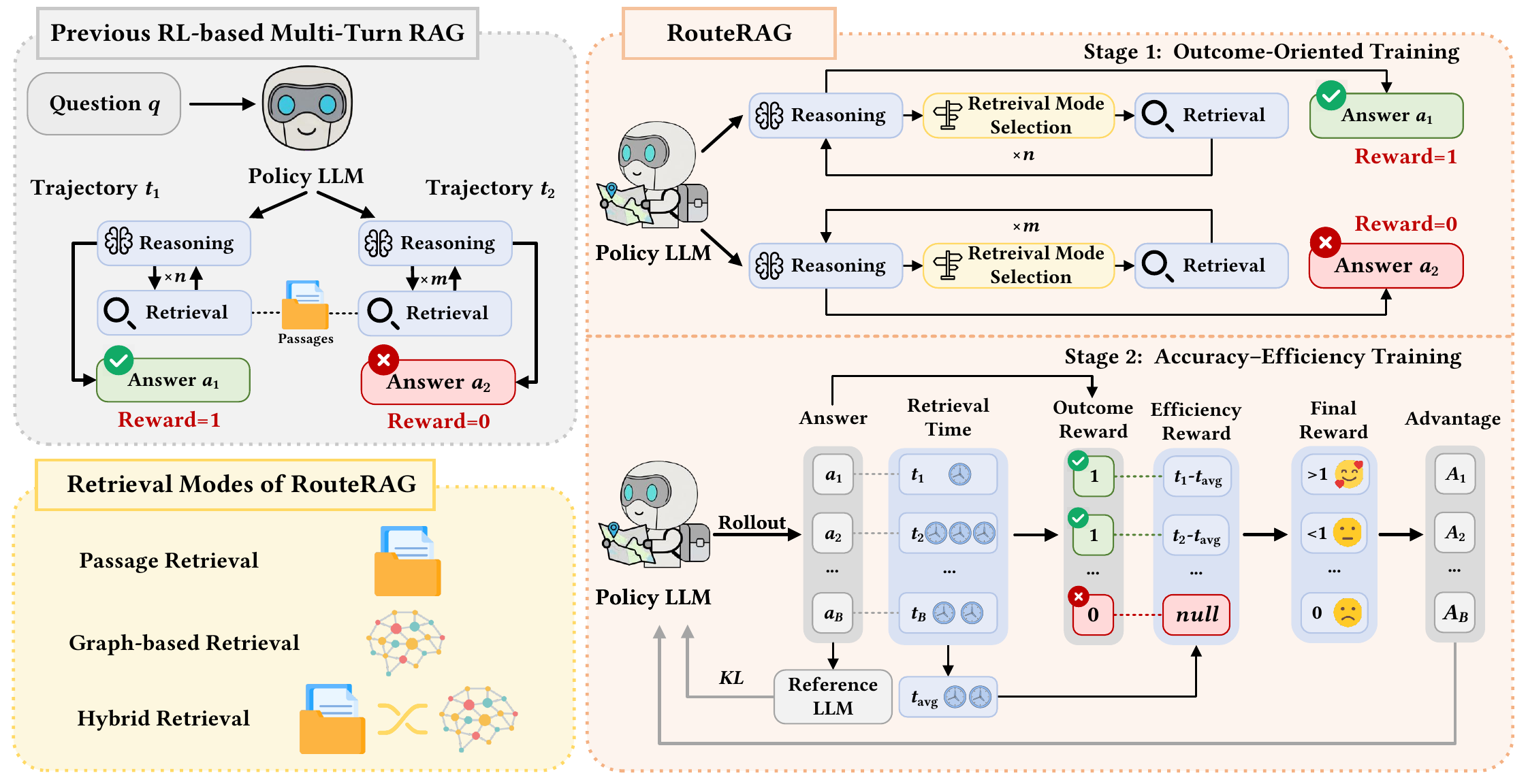}
    \caption{Previous RL-based multi-turn RAG vs. \model{}. Prior methods mainly focus on interleaving reasoning with passage retrieval and reward on answer correctness. \model{} extends retrieval to passage, graph, and hybrid modes, and is trained with a two-stage RL framework that optimizes both accuracy and efficiency.}
    \label{fig:method}
\end{figure*}
\subsubsection{Hybrid Knowledge Access}
\label{sec:retriever}
In \model{}, the retriever $\mathcal{R}$ is responsible for providing external knowledge to support reasoning, with three different retrieval modes. 

\noindent \textbf{Passage Retrieval}. The passage retriever is implemented with Dense Passage Retrieval (DPR)~\citep{Karpukhin2020dense}, which encodes both the sub-query and all passages in the corpus into a shared embedding space. Retrieval is performed by computing similarity scores between the query vector and passage vectors, and the top-$k$ passages are selected as evidence.  

\noindent \textbf{Graph-based Retrieval}. The graph retriever is implemented based on HippoRAG 2~\citep{Gutierrez2025from}, which first constructs a knowledge graph over passages. Given a sub-query, the retriever applies personalized PageRank over the graph to propagate relevance from query-linked nodes, thereby identifying passages that are related to the query through multi-hop connections.  

\noindent \textbf{Hybrid Retrieval}. The hybrid retriever combines passage and graph retrieval using Reciprocal Rank Fusion (RRF)~\citep{Cormack2009reciprocal}. Specifically, given two ranked lists, each document is assigned a fused score that decreases with its reciprocal rank in each list, which ensures that documents highly ranked by either retrieval mode are promoted in the merged list. Formally, the fused score is defined as
\begin{equation}
    \text{RRF}(d) = \sum_{m \in \{\texttt{Passage}, \texttt{Graph}\}} \frac{1}{k + \text{rank}_m(d)},
\end{equation}
where $\text{rank}_m(d)$ denotes the rank position of document $d$ in retrieval mode $m$, and $k$ is a smoothing hyperparameter. Documents are then re-ranked according to $\text{RRF}(d)$ to form the final hybrid list.

\subsection{Two-Stage Reinforcement Learning}
\label{sec:two_stage_RL}
To optimize the unified generation policy, \model{} is trained with a two-stage RL framework based on GRPO. The motivation is to first ensure that the model acquires the basic ability to produce correct answers, and then to further refine its retrieval strategy to improve efficiency without sacrificing accuracy, as shown in \Cref{fig:method}. 
In this section, we introduce the reward design that guides the learning objectives (\Cref{sec:reward_design}) and the training algorithm that realizes the optimization procedure (\Cref{sec:training_algorithm}).

\subsubsection{Reward Design}
\label{sec:reward_design}
RL optimization is fundamentally guided by the reward signal. To support the two-stage training, we devise different rewards for each stage, i.e., outcome-oriented reward and accuracy--efficiency reward.

\noindent \textbf{Stage 1: Outcome-Oriented Reward.} In the first stage, the reward is defined purely by the correctness of the model output. Specifically, the reward is set to 1 if the generated answer $y$ exactly matches the ground-truth label $y^*$, and 0 otherwise:
\begin{equation}
    R_{\phi}(x, y) = \text{EM}(y, y^*).
\end{equation}

\noindent \textbf{Stage 2: Accuracy--Efficiency Reward.}
In the second stage, we extend the reward function to jointly optimize for correctness and retrieval efficiency. The reward is defined as
\begin{equation}
R_{\phi}(x,y) =
\begin{cases}
R_{\text{outcome}},\!\!&\!\!R_{\text{outcome}}=0\\
R_{\text{outcome}} + R_{\text{efficiency}},\!\!&\!\!R_{\text{outcome}}=1
\end{cases},
\end{equation}
where $R_{\text{outcome}} \in \{0,1\}$ denotes exact match accuracy. The efficiency reward $R_{\text{efficiency}}$ is computed from the total retrieval time across all reasoning steps, and only for those trajectories that correctly reach the answer. We apply a centered scaling by subtracting the average retrieval time $t_{\text{avg}}$, such that
\begin{equation}
    R_{\text{efficiency}} = \frac{t_{\text{avg}}-t}{T},
\end{equation}
where $t$ is the total retrieval time for the current trajectory, $t_{\text{avg}}$ is the average retrieval time of the current batch, and $T$ is a normalization constant ensuring the value of $t$ and $t_{\text{avg}}$ $\in[0,0.5]$.
This design provides positive reward for trajectories that achieve the correct answer at a pace exceeding the average, while imposing penalties on those that do not, thereby encouraging the model to retrieve more selectively without sacrificing answer quality.

\subsubsection{Training Algorithm}
\label{sec:training_algorithm}
We adopt GRPO~\citep{Shao2024deepseekmath, Guo2025deepseek} to train the unified generation policy $\pi_\theta$ over interleaved reasoning and retrieval actions. 
GRPO stabilizes learning by comparing trajectories within a group, thereby reducing variance in sparse-reward settings.

The policy model $\pi_\theta$ is optimized by maximizing the following objective:
\begin{equation}
\small
\label{eq:grpo_objective}
\begin{aligned}
J_{\text{GRPO}}(\theta) = & \mathbb{E}_{x \sim Q, \{y_i\}_{i=1}^G \sim \pi_{\theta_{\text{old}}}(Y \mid q)}\frac{1}{G}\sum_{i=1}^{G} \\ &\Bigg[ \min \Big(r_i(\theta) A_i, {\text{clip}}(r_i(\theta), 1-\epsilon, 1+\epsilon) A_i\Big) \\ & - \beta \, \mathbb{D}_{\text{KL}}[\pi_{\theta_{\text{old}}} \| \pi_\theta]\Bigg],
\end{aligned}
\end{equation}
where $\epsilon$ and $\beta$ are hyperparameters, $\pi_{\theta_\text{old}}$ denotes the old policy, $r_i(\theta) = \frac{\pi_\theta(y_i \mid x)}{\pi_{\theta_\text{old}}(y_i \mid x)}$, $A_i$ denotes the group-relative advantage for the $i$-th trajectory, and the KL penalty $\mathbb{D}_\text{KL}[\pi_{\theta_\text{old}} \| \pi_\theta]$ regularizes the new policy against deviating excessively from the old policy.
Further theoretical analysis on the effectiveness of the efficiency reward and GRPO is provided in~\Cref{appendix:efficiency_reward_analysis}.

\section{Experiments}
\subsection{Experimental Setting}
\textbf{Evaluation Datasets.}
Following~\citet{Jimenez2024hipporag} and~\citet{Gutierrez2025from}, we evaluate \model{} on five widely used benchmarks for simple and multi-hop QA, namely PopQA~\citep{Mallen2023popqa}, Natural Questions (NQ)~\citep{Kwiatkowski2019natural, Wang2024rear}, HotpotQA~\citep{Yang2018hotpotqa}, 2WikiMultihopQA (2Wiki)~\citep{Ho2020constructing}, and MuSiQue~\citep{Trivedi2022musique}. PopQA is an open-domain QA dataset designed to evaluate factual recall over long-tail knowledge, and NQ contains naturally occurring queries paired with answers from Wikipedia. HotpotQA and 2Wiki focus on multi-hop reasoning across Wikipedia passages, while MuSiQue requires reasoning over compositional sub-questions. 

\begin{table*}[t]
\centering
\small
\renewcommand{\arraystretch}{1.2}
\resizebox{\textwidth}{!}{%
\begin{tabular}{lcccccccccccc}
\toprule
\multirow{3}{*}{\textbf{Method}} 
& \multicolumn{4}{c}{\textbf{Simple QA}} 
& \multicolumn{6}{c}{\textbf{Multi-hop QA}} 
& \multicolumn{2}{c}{\multirow{2}{*}{\textbf{Average}}}\\
\cmidrule(lr){2-5} \cmidrule(lr){6-11}
 & \multicolumn{2}{c}{\textbf{PopQA}} 
 & \multicolumn{2}{c}{\textbf{NQ}} 
 & \multicolumn{2}{c}{\textbf{HotpotQA}} 
 & \multicolumn{2}{c}{\textbf{2Wiki}} 
 & \multicolumn{2}{c}{\textbf{MuSiQue}}
 & & \\
\cmidrule(lr){2-3} \cmidrule(lr){4-5} \cmidrule(lr){6-7} \cmidrule(lr){8-9} \cmidrule(lr){10-11} \cmidrule(lr){12-13}
 & \textbf{EM} & \textbf{F1} 
 & \textbf{EM} & \textbf{F1} 
 & \textbf{EM} & \textbf{F1} 
 & \textbf{EM} & \textbf{F1} 
 & \textbf{EM} & \textbf{F1}
 & \textbf{EM} & \textbf{F1}\\
\midrule
\rowcolor{gray!10}
\multicolumn{13}{c}{{\centering\raisebox{-.3\height}{\includegraphics[width=0.4cm]{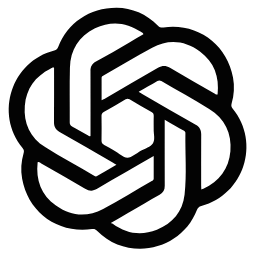}}} \textit{\textbf{GPT-4o-mini}}}\\
\midrule
Direct Inference & 16.1 & 22.7 & 35.2 & 52.7 & 28.6 & 41.0 & 30.2 & 36.3 & 11.2 & 22.0 & 24.3 & 34.9 \\
\multicolumn{13}{l}{\textit{\textbf{Graph-based RAG}}}\\
GraphRAG  & 30.7 & 51.3 & \underline{38.0} & \underline{55.5} & \underline{51.4} & \underline{67.6} & 45.7 & 61.0 & 27.0 & \underline{42.0} & 38.6 & \underline{55.5} \\
LightRAG   & 1.9 & 14.8 & 2.8 & 15.4 & 9.9 & 20.2 & 2.5 & 12.1 & 2.0 & 9.3 & 3.8 & 14.4 \\
RAPTOR  & \underline{41.9} & 55.1 & 37.8 & 54.5 & 50.6 & 64.7 & 39.7 & 48.4 & \underline{27.7} & 39.2 & 39.5 & 52.4 \\
HippoRAG   & \textbf{42.5} & \textbf{56.2} & 37.2 & 52.5 & 46.3 & 60.0 & \underline{59.4} & \underline{67.3} & 24.0 & 35.9 & \underline{41.9} & 54.4 \\
HippoRAG 2 & 41.7 & \underline{55.7} & \textbf{43.4} & \textbf{60.0} & \textbf{56.3} & \textbf{71.1} & \textbf{60.5} & \textbf{69.7} & \textbf{35.0} & \textbf{49.3} & \textbf{47.4} & \textbf{61.2} \\
\midrule
\rowcolor{gray!10}
\multicolumn{13}{c}{{\centering\raisebox{-.3\height}{\includegraphics[width=0.4cm]{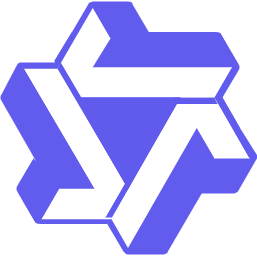}}} \textit{\textbf{Qwen2.5-3B}}}\\
\midrule
Vanilla RAG & 30.3 & 41.6 & 18.1 & 31.8 & 29.5 & 41.8 & 19.7 & 27.4 & 10.3 & 17.5 & 21.6 & 32.0 \\
\multicolumn{13}{l}{\textit{\textbf{Graph-based RAG}}}\\
HippoRAG 2 & 29.1 & 40.1 & 20.3 & 33.5 & 31.2 & 45.0 & 21.5 & 33.8 & 12.2 & 20.2 & 22.9 & 34.5 \\
\multicolumn{13}{l}{\textit{\textbf{Multi-turn RAG}}}\\
Search-o1 & 17.1 & 23.8 & 19.9 & 29.1 & 18.7 & 26.3 & 16.9 & 20.9 & 3.9 & 10.5 & 15.3 & 22.1 \\
Search-R1 & \underline{45.8} & \underline{53.3} & \textbf{46.2}$^*$ & \textbf{54.8}$^*$ & \underline{45.2}$^*$ & \underline{56.9}$^*$ & \underline{42.4} & \underline{50.8} & \underline{22.2} & \underline{30.9} & \underline{40.4} & \underline{49.3} \\
\rowcolor{lavender}
\textbf{\model{}-3B} & \textbf{49.4} & \textbf{56.8} & \underline{44.1} & \underline{53.4} & \textbf{53.2}$^*$ & \textbf{65.1}$^*$ & \textbf{57.5} & \textbf{64.1} & \textbf{30.7} & \textbf{39.3} & \textbf{47.0} & \textbf{55.7} \\
\midrule
\rowcolor{gray!10}
\multicolumn{13}{c}{{\centering\raisebox{-.3\height}{\includegraphics[width=0.4cm]{figs/logo/qwen.png}}} \textit{\textbf{Qwen2.5-7B}}}\\
\midrule
Vanilla RAG & 26.3 & 37.8 & 11.0 & 30.6 & 29.1 & 47.5 & 17.2 & 33.3 & 11.4 & 23.0 & 19.0 & 34.4 \\
\multicolumn{13}{l}{\textit{\textbf{Graph-based RAG}}}\\
HippoRAG 2 & 27.0 & 37.9 & 8.1 & 27.2 & 27.4 & 46.1 & 16.8 & 34.7 & 12.3 & 24.0 & 18.3 & 34.0 \\
\multicolumn{13}{l}{\textit{\textbf{Multi-turn RAG}}}\\
Search-o1 & 4.7 & 7.2 & 18.1 & 27.5 & 13.5 & 19.1 & 6.4 & 7.9 & 2.9 & 7.7 & 9.1 & 13.9 \\
R1-Searcher & 28.4 & 41.0 & 41.6 & 52.2 & 46.6$^*$ & 56.7$^*$ & 41.7$^*$ & 49.0$^*$ & 29.3 & 37.6 & 37.5 & 47.3 \\
Search-R1 & \textbf{51.3} & \textbf{57.1} & \textbf{56.8}$^*$ & \textbf{65.3}$^*$ & \underline{51.0}$^*$ & \underline{62.0}$^*$ & \underline{51.8} & \underline{58.9} & \underline{32.0} & \underline{40.8} & \underline{48.6} & \underline{56.8} \\
\rowcolor{lavender}
\textbf{\model{}-7B} & \underline{50.6} & \underline{56.4} & \underline{51.5} & \underline{60.4} & \textbf{60.8}$^*$ & \textbf{72.5}$^*$ & \textbf{57.1} & \textbf{64.6} & \textbf{39.6} & \textbf{49.3} & \textbf{51.9} & \textbf{60.6} \\
\bottomrule
\end{tabular}}
\caption{Main results on simple and multi-hop QA benchmarks. The best results within each backbone group are indicated in bold, while the underlined values represent the second-best results. $^*$ represents in-domain datasets.}
\label{tab:main_results}
\end{table*}

\noindent \textbf{Baselines.}
We compare \model{} against several types of representative approaches:  
(1) Vanilla RAG~\citep{Lewis2020retrieval}, which performs single-shot dense passage retrieval and generation.  
(2) Multi-turn RAG methods, including Search-o1~\citep{Li2025search}, Search-R1~\citep{Jin2025search}, and R1-Searcher~\citep{Song2025r1}, wherein the latter two methods utilize RL to enhance multi-turn passage RAG.
(3) Graph-based RAG methods, including GraphRAG~\citep{Edge2024local}, LightRAG~\citep{Guo2024lightrag}, RAPTOR~\citep{Sarthi2024raptor}, HippoRAG~\citep{Jimenez2024hipporag}, and HippoRAG 2~\citep{Gutierrez2025from}, which leverage structured knowledge graphs for retrieval. 

\noindent \textbf{Implementation Details.}
We conduct training using Qwen2.5-3B-Instruct and Qwen2.5-7B-Instruct~\citep{Qwen2025qwen25} as the backbone models. The training data consists of 10k sampled queries from the HotpotQA training set~\citep{Yang2018hotpotqa}, while the retrieval corpus is built from their associated documents. For retrieval, we adopt Contriever~\citep{Izacard2022unsupervised} and NV-Embed-v2~\citep{Lee2025nvembed} as the dense retriever for 3B and 7B models, respectively.
For baseline evaluations, text-based RAG systems are assessed under the same Qwen2.5 backbone, while graph-based RAG systems utilize the GPT-4o-mini backbone. In particular, HippoRAG 2~\citep{Gutierrez2025from}, the strongest graph-based baseline, is evaluated employing both Qwen2.5 and GPT-4o-mini backbones. 
We report Exact Match (EM) and F1 scores as evaluation metrics. Additional implementation details, including the training prompt template, hyperparameters, and training configuration, are provided in~\Cref{appendix:implementation_details}.

\subsection{Main Results}
We conduct a comprehensive comparison of \model{} against all the baseline methods, as shown in~\Cref{tab:main_results}. From the results, we make the following key observations:

\textbf{(1) \model{} substantially improves the performance of a small backbone, especially on multi-hop QA.}
Graph-based methods such as HippoRAG 2 perform well with the strong GPT-4o-mini backbone but drop sharply with the smaller Qwen2.5-3B and Qwen2.5-7B models, indicating that small LLMs struggle to handle complex reasoning chains. In contrast, \model{} achieves much better performance on small backbones by jointly learning reasoning, retrieval, and answer generation within a unified policy model.

\begin{table*}[t]
\centering
\small
\renewcommand{\arraystretch}{1.2}
\resizebox{\textwidth}{!}{%
\begin{tabular}{lcccccccccccc}
\toprule
\multirow{2}{*}{\textbf{Method}} 
 & \multicolumn{2}{c}{\textbf{PopQA}} 
 & \multicolumn{2}{c}{\textbf{NQ}} 
 & \multicolumn{2}{c}{\textbf{HotpotQA}} 
 & \multicolumn{2}{c}{\textbf{2Wiki}} 
 & \multicolumn{2}{c}{\textbf{MuSiQue}}
 & \multicolumn{2}{c}{\textbf{Average}}\\
\cmidrule(lr){2-3} \cmidrule(lr){4-5} \cmidrule(lr){6-7} \cmidrule(lr){8-9} \cmidrule(lr){10-11} \cmidrule(lr){12-13}
 & \textbf{EM} & \textbf{F1} 
 & \textbf{EM} & \textbf{F1} 
 & \textbf{EM} & \textbf{F1} 
 & \textbf{EM} & \textbf{F1} 
 & \textbf{EM} & \textbf{F1}
 & \textbf{EM} & \textbf{F1}\\
\midrule
\textbf{\model{}-3B} & \textbf{49.4} & \textbf{56.8} & \textbf{44.1} & \textbf{53.4} & 53.2 & \underline{65.1} & \textbf{57.5} & \underline{64.1} & \textbf{30.7} & \textbf{39.3} & \textbf{47.0} & \textbf{55.7} \\
\multicolumn{1}{r}{\textit{w/o} Stage 2 training} & 47.0 & 54.6 & 39.1 & 49.4 & 46.4 & 58.6 & 48.5 & 55.6 & 24.3 & 33.0 & 41.1 & 50.2 \\
\multicolumn{1}{r}{\textit{w/o} training} & 26.2 & 41.3 & 18.6 & 30.5 & 35.4 & 46.9 & 24.4 & 37.9 & 15.4 & 24.4 & 24.0 & 36.2 \\
\midrule
\multicolumn{1}{r}{\textit{w/ only} passage retrieval} & 48.7 & 55.2 & \underline{43.9} & \underline{53.3} & 53.1 & 64.5 & 53.0 & 58.9 & 28.0 & 36.2 & 45.3 & 53.6 \\
\multicolumn{1}{r}{\textit{w/ only} graph-based retrieval} & \underline{49.3} & \textbf{56.8} & 43.8 & 53.0 & \textbf{53.4} & \textbf{65.5} & \underline{57.4} & \textbf{64.2} & \underline{29.7} & \underline{38.3} & \underline{46.7} & \underline{55.6} \\
\multicolumn{1}{r}{\textit{w/ only} hybrid retrieval} & 48.7 & \underline{55.9} & 43.1 & 52.5 & \underline{53.3} & \underline{65.1} & 53.7 & 59.8 & 28.8 & 37.0 & 45.5 & 54.0 \\
\bottomrule\end{tabular}}
\caption{Ablation studies on RL training and hybrid retrieval. ``\textit{w/o} training'' denotes the base model without any RL training, and ``\textit{w/o} stage 2 training'' denotes training only with the first stage. For hybrid retrieval ablation, we compare \model{} with variants restricted to only passage retrieval, only graph retrieval, or only hybrid retrieval.}
\label{tab:ablation}
\end{table*}
\begin{table*}[t]
\centering
\small
\renewcommand{\arraystretch}{1.2}
\resizebox{\textwidth}{!}{%
\begin{tabular}{lcccccccccccc}
\toprule
\multirow{2}{*}{\textbf{Method}} 
 & \multicolumn{2}{c}{\textbf{PopQA}} 
 & \multicolumn{2}{c}{\textbf{NQ}} 
 & \multicolumn{2}{c}{\textbf{HotpotQA}} 
 & \multicolumn{2}{c}{\textbf{2Wiki}} 
 & \multicolumn{2}{c}{\textbf{MuSiQue}}
 & \multicolumn{2}{c}{\textbf{Average}}\\
\cmidrule(lr){2-3} \cmidrule(lr){4-5} \cmidrule(lr){6-7} \cmidrule(lr){8-9} \cmidrule(lr){10-11} \cmidrule(lr){12-13}
 & \textbf{EM} & \textbf{F1} 
 & \textbf{EM} & \textbf{F1} 
 & \textbf{EM} & \textbf{F1} 
 & \textbf{EM} & \textbf{F1} 
 & \textbf{EM} & \textbf{F1}
 & \textbf{EM} & \textbf{F1}\\
\midrule
\textbf{\model{}-3B} & \textbf{49.4} & \textbf{56.8} & \textbf{44.1} & \textbf{53.4} & 53.2 & \underline{65.1} & \textbf{57.5} & \underline{64.1} & \textbf{30.7} & \textbf{39.3} & \textbf{47.0} & \textbf{55.7} \\
\multicolumn{1}{r}{\textit{w/o} efficiency reward} & \underline{41.5} & \underline{54.1} & \underline{41.1} & \underline{51.7} & \textbf{53.7} & \textbf{66.2} & \underline{56.9} & \textbf{65.0} & \underline{30.2} & \underline{38.9} & \underline{44.7} & \underline{55.2} \\
\midrule
\textbf{\model{}-7B} & \textbf{50.6} & \textbf{56.4} & \underline{51.5} & \underline{60.4} & \textbf{60.8} & \underline{72.5} & \textbf{57.1} & \textbf{64.6} & \underline{39.6} & \underline{49.3} & \textbf{51.9} & \textbf{60.6} \\
\multicolumn{1}{r}{\textit{w/o} efficiency reward} & \underline{46.0} & \underline{54.1} & \textbf{52.5} & \textbf{62.1} & \underline{59.9} & \textbf{72.6} & \underline{49.5} & \underline{56.8} & \textbf{42.4} & \textbf{52.0} & \underline{50.1} & \underline{59.5} \\
\bottomrule\end{tabular}}
\caption{Ablation study on efficiency reward. ``w/o efficiency reward'' denotes training only with EM-based outcome rewards.}
\label{tab:ablation_efficiency}
\end{table*}

\textbf{(2) \model{} approaches GPT-4o-mini-based graph-based RAG systems despite using a much smaller model.}
Despite the large performance gap usually observed between GPT-4o-mini and Qwen2.5-3B/7B, \model{} narrows this gap substantially and even surpasses several graph-based systems built on GPT-4o-mini. This suggests that improving the policy can be as impactful as scaling up the backbone itself. 

\textbf{(3) \model{} outperforms the strongest RL-trained multi-turn baseline with much smaller training cost.}
Search-R1, the prior strongest RL-based multi-turn system, is trained on 170k questions from NQ and HotpotQA. 
Despite being trained on a mere 10k HotpotQA instances, \model{} achieves the best average scores among all 3B and 7B methods, demonstrating that structured retrieval and retrieval mode selection can yield more effective and sample-efficient multi-turn RAG policies than scaling training data alone. While \model{} is slightly weaker on simple QA, this is expected because the training data is dominated by multi-hop questions. Nevertheless, its overall performance remains competitive given the smaller training cost.

\subsection{Ablation Study}
In this section, we conduct ablation experiments to validate the effectiveness of RL training, hybrid retrieval, and the efficiency reward.

\noindent \textbf{RL Training.}
The upper part of Table~\ref{tab:ablation} compares the full model against two ablated variants: (1) a model trained only with the first stage, and (2) the untrained backbone.
The results show that \model{} already acquires a strong reasoning ability through the Stage 1 training, outperforming the untrained backbone by a large margin. The Stage 2 training further improves the performance of \model{}, especially on multi-hop QA datasets.

\noindent \textbf{Hybrid Retrieval.}
The lower part of Table~\ref{tab:ablation} presents an ablation study comparing \model{} with variants restricted to a single retrieval mode. Passage retrieval performs well on simple QA, while graph retrieval is more effective for multi-hop reasoning, confirming their complementary strengths. The full \model{} achieves the highest average accuracy by dynamically selecting among the three modes, showing that adaptive retrieval yields stronger robustness and generalization than any fixed strategy.

\begin{figure*}[thbp]
\centering
\begin{subfigure}{0.48\textwidth}
    \includegraphics[width=\textwidth]{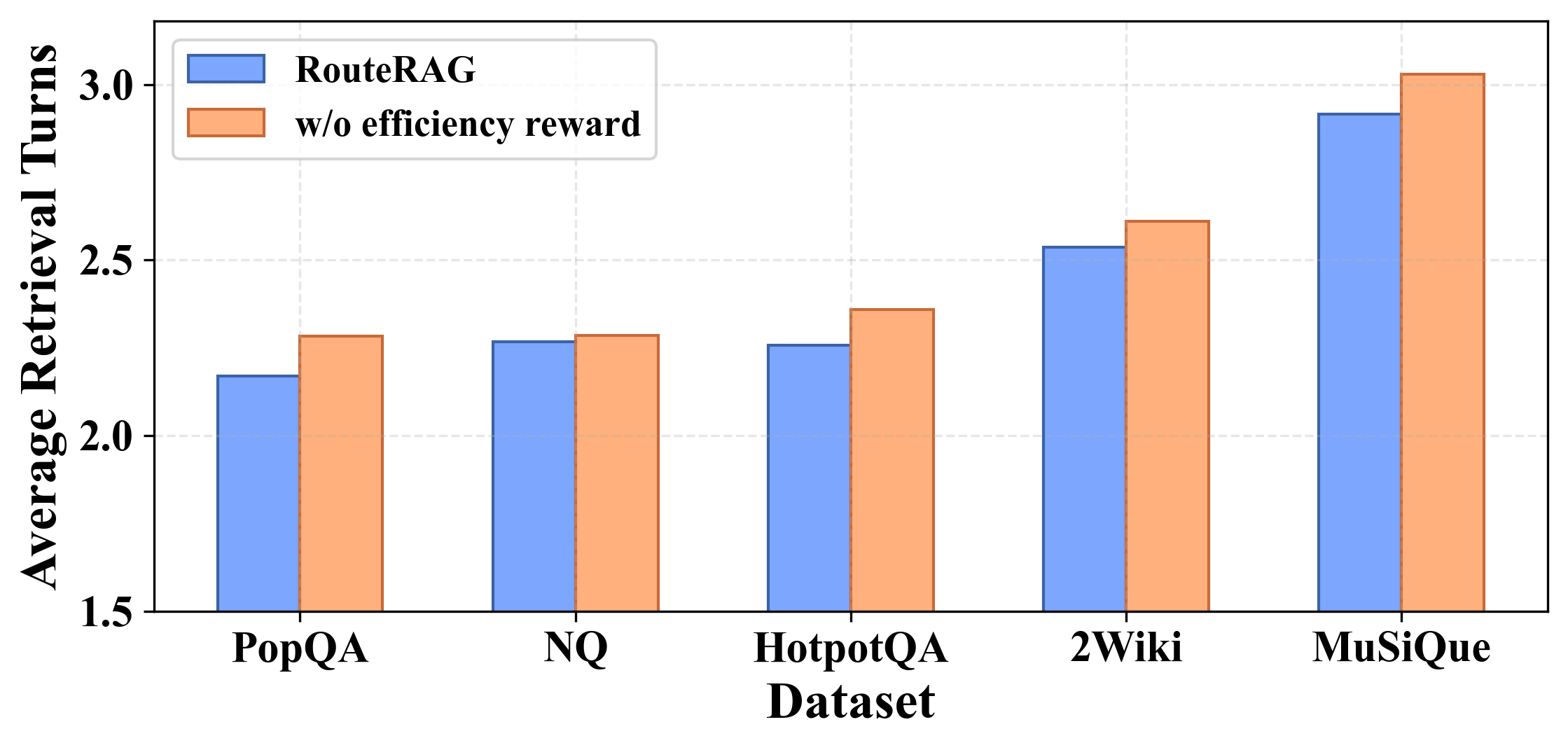}
    \caption{\model{}-3B.}
    \label{fig:retrieval_turns_3B}
\end{subfigure}
\hfill
\begin{subfigure}{0.48\textwidth}
    \includegraphics[width=\textwidth]{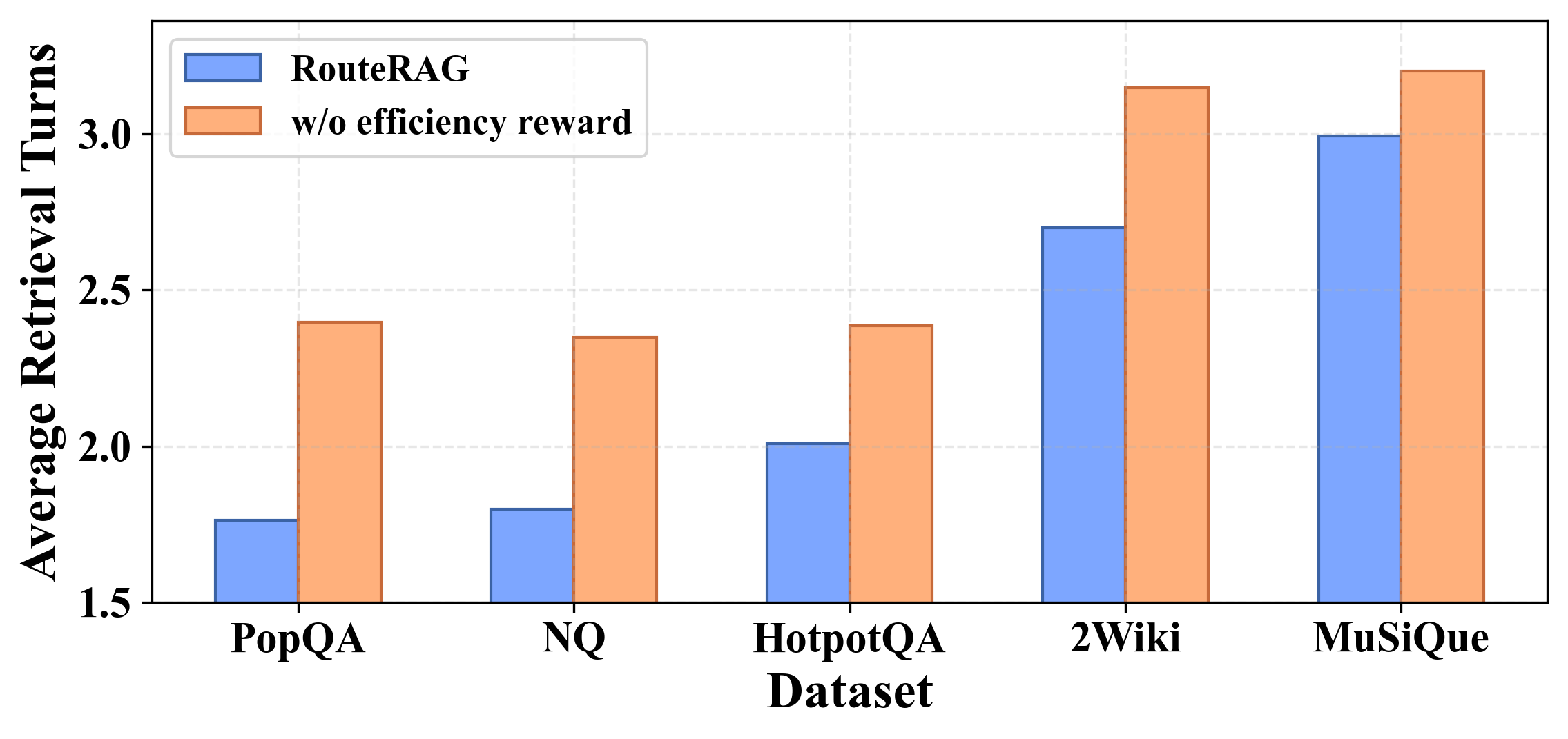}
    \caption{\model{}-7B.}
    \label{fig:retrieval_turns_7B}
\end{subfigure}
\caption{Comparing the average retrieval turns of \model{} and its variant without efficiency reward.}
\label{fig:efficiency_turns}
\end{figure*}
\noindent \textbf{Efficiency Reward.}
We evaluate whether \model{} learns to retrieve evidence more efficiently while keeping effectiveness with the efficiency reward by comparing it to a variant trained only with the outcome reward, using the same number of training steps.
As shown in \Cref{tab:ablation_efficiency}, \model{} maintains comparable or even higher accuracy than its outcome rewards-only counterpart.
\Cref{fig:efficiency_turns} shows that both \model{}-3B and \model{}-7B consistently reduce the average retrieval turns across all datasets, with larger savings observed on the 7B model. 
\Cref{tab:efficiency_accuracy} further assesses the trade-off between retrieval turns and task accuracy, highlighting that efficiency gains do not come at the cost of answer quality. 
This demonstrates that incorporating efficiency rewards encourages the policy to avoid unnecessary retrieval steps while still collecting sufficient evidence.
A comprehensive analysis of this phenomenon is available in \Cref{appendix:selective_retrieval}.
\begin{table}[tbp]
\centering
\small
\caption{Average retrieval turns and accuracy of \model{}.}
\label{tab:efficiency_accuracy}
\begin{tabular}{lccc}
\toprule
\textbf{Method} & \textbf{Avg. Turns} & \textbf{Avg. F1} \\
\midrule
\textbf{\model{}-3B} & 2.43 & 55.7 \\
\multicolumn{1}{r}{\textit{w/o} efficiency reward} & 2.51 \textcolor{brickred}{(+3.3\%)} & 55.2 \textcolor{brickred}{(-0.5)} \\
\midrule
\textbf{\model{}-7B} & 2.25 & 60.6 \\
\multicolumn{1}{r}{\textit{w/o} efficiency reward} & 2.70 \textcolor{brickred}{(+20\%)} & 59.5 \textcolor{brickred}{(-1.1)} \\
\bottomrule
\end{tabular}
\end{table}

\subsection{Analysis on Reasoning Ability}
\begin{figure}
    \centering
    \includegraphics[width=\linewidth]{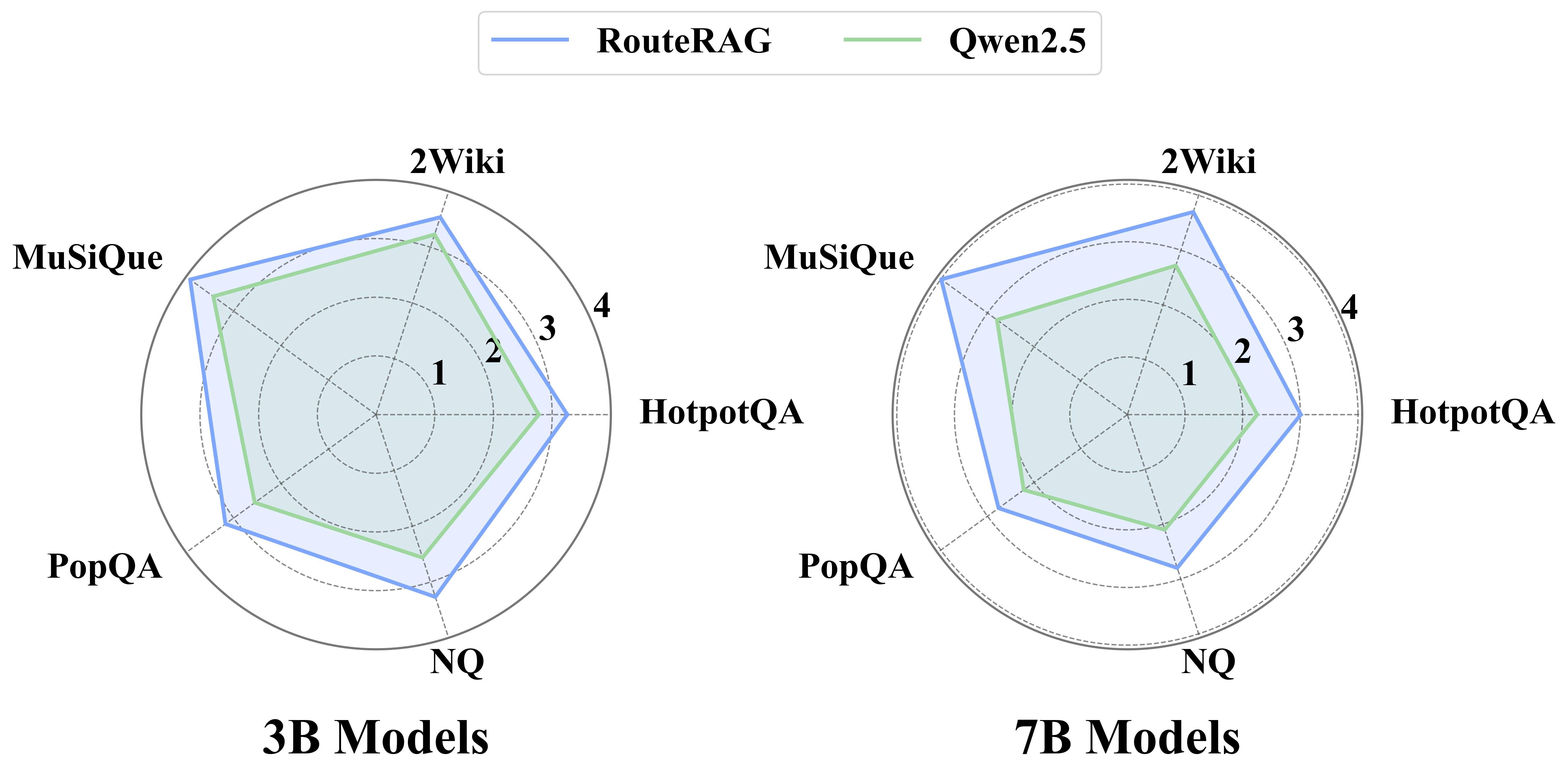}
    \caption{Comparison of average reasoning steps.}
    \label{fig:reasoning_turns}
\end{figure}

\begin{figure}
    \centering
    \includegraphics[width=\linewidth]{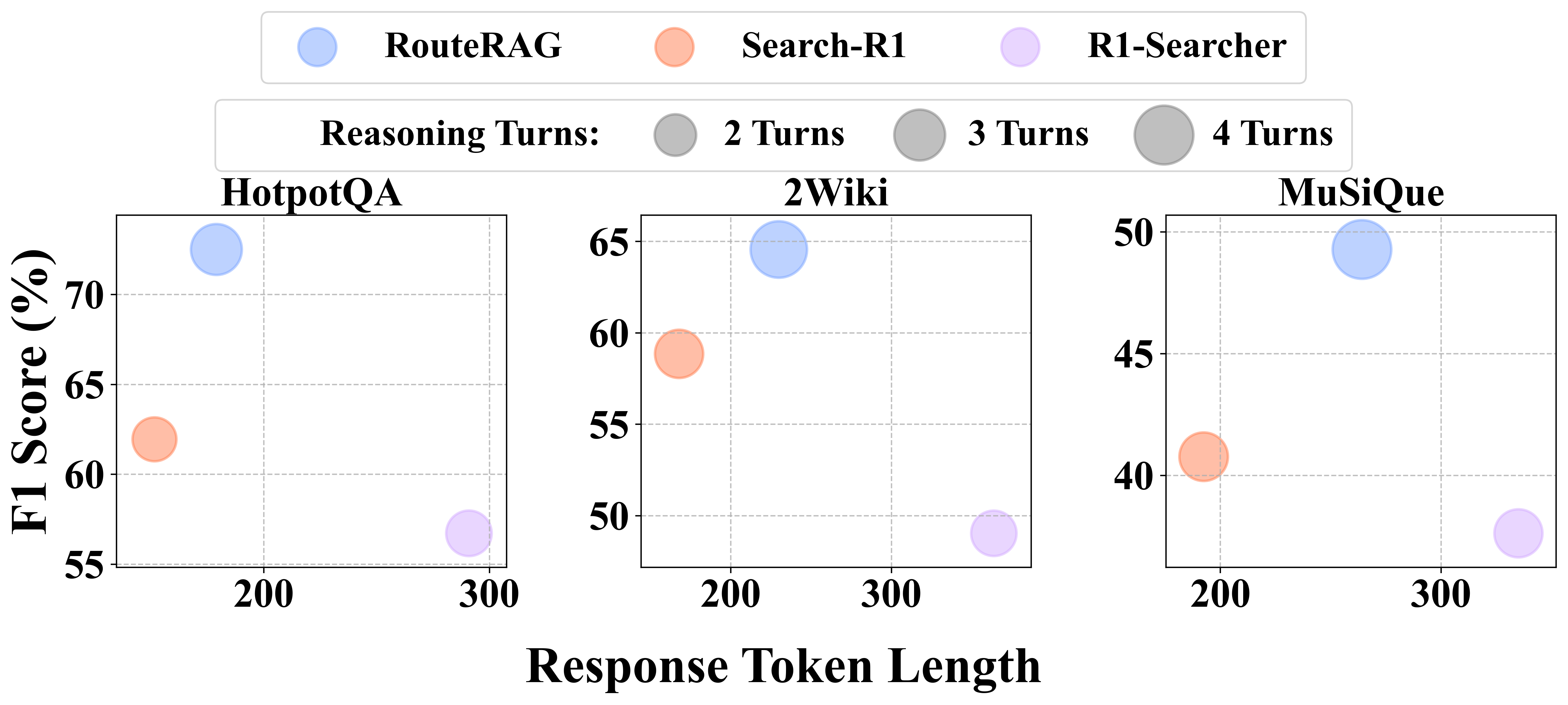}
    \caption{Comparison in terms of performance, response token length, and reasoning turns.}
    \label{fig:reasoning_tokens}
\end{figure}

To better understand how the multi-round reasoning ability of \model{} evolves during training, we analyze the average number of reasoning steps before and after training. 
As shown in~\Cref{fig:reasoning_turns}, both the 3B and 7B models exhibit increased reasoning depth after training. Compared with \model{}-3B, \model{}-7B shows a more pronounced improvement, with larger gains on the more complex MuSiQue and 2Wiki datasets, and smaller increases on PopQA and NQ.
\Cref{fig:reasoning_tokens} further expands the analysis by jointly examining performance, response token length, and reasoning turns across datasets. The results show that \model{} consistently achieves a more favorable balance among the three factors compared with Search-R1 and R1-Searcher, achieving high F1 with moderate token length.
For a qualitative view of how the reasoning behavior changed in practice, case studies comparing model outputs before and after training are presented in~\Cref{appendix:case_study}.

\section{Conclusions}
In this paper, we presented \model{}, an RL framework for efficient multi-turn hybrid RAG. Unlike prior multi-turn RAG systems that rely on static prompting or single-mode retrieval, our approach learns a unified policy that interleaves reasoning, retrieval mode selection, retrieval query generation, and answer generation.
Our two-stage training framework further ensures that the model first acquires robust answer correctness and then improves retrieval efficiency without sacrificing accuracy. Experiments conducted on five knowledge-intensive QA benchmarks demonstrate that \model{} significantly outperforms existing graph-based and multi-turn RAG systems, highlighting that efficiency gains can be achieved without compromising answer quality. 

\section*{Limitations}
Despite the strong empirical results, our proposed \model{} has several limitations. 
First, due to computational constraints, we conduct RL training and evaluation only on 3B and 7B LLMs. Larger or more diverse model architectures may exhibit different behaviors under our training framework. 
Second, our experiments utilize HippoRAG 2 as the graph retriever. While we adopt it because it is currently the strongest graph-based RAG system, this choice limits our evaluation of how \model{} interacts with alternative graph retrievers or graph construction pipelines.

\bibliography{custom}

\begin{thebibliography}{45}
\providecommand{\natexlab}[1]{#1}

\bibitem[{Achiam et~al.(2023)Achiam, Adler, Agarwal, Ahmad, Akkaya, Aleman,
  Almeida, Altenschmidt, Altman, Anadkat et~al.}]{Achiam2023gpt}
Josh Achiam, Steven Adler, Sandhini Agarwal, Lama Ahmad, Ilge Akkaya,
  Florencia~Leoni Aleman, Diogo Almeida, Janko Altenschmidt, Sam Altman,
  Shyamal Anadkat, and 1 others. 2023.
\newblock \href {https://arxiv.org/abs/2303.08774} {Gpt-4 technical report}.
\newblock \emph{arXiv preprint arXiv:2303.08774}.

\bibitem[{Augenstein et~al.(2024)Augenstein, Baldwin, Cha, Chakraborty,
  Ciampaglia, Corney, DiResta, Ferrara, Hale, Halevy
  et~al.}]{Augenstein2024factuality}
Isabelle Augenstein, Timothy Baldwin, Meeyoung Cha, Tanmoy Chakraborty,
  Giovanni~Luca Ciampaglia, David Corney, Renee DiResta, Emilio Ferrara, Scott
  Hale, Alon Halevy, and 1 others. 2024.
\newblock \href {https://doi.org/10.1038/s42256-024-00881-z} {Factuality
  challenges in the era of large language models and opportunities for
  fact-checking}.
\newblock \emph{Nature Machine Intelligence}, 6(8):852--863.

\bibitem[{Christiano et~al.(2017)Christiano, Leike, Brown, Martic, Legg, and
  Amodei}]{Christiano2017deep}
Paul~F Christiano, Jan Leike, Tom Brown, Miljan Martic, Shane Legg, and Dario
  Amodei. 2017.
\newblock \href
  {https://proceedings.neurips.cc/paper_files/paper/2017/file/d5e2c0adad503c91f91df240d0cd4e49-Paper.pdf}
  {Deep reinforcement learning from human preferences}.
\newblock \emph{Advances in neural information processing systems}, 30.

\bibitem[{Cormack et~al.(2009)Cormack, Clarke, and
  Buettcher}]{Cormack2009reciprocal}
Gordon~V Cormack, Charles~LA Clarke, and Stefan Buettcher. 2009.
\newblock \href {https://doi.org/10.1145/1571941.1572114} {Reciprocal rank
  fusion outperforms condorcet and individual rank learning methods}.
\newblock In \emph{Proceedings of the 32nd international ACM SIGIR conference
  on Research and development in information retrieval}, pages 758--759.

\bibitem[{Edge et~al.(2024)Edge, Trinh, Cheng, Bradley, Chao, Mody, Truitt,
  Metropolitansky, Ness, and Larson}]{Edge2024local}
Darren Edge, Ha~Trinh, Newman Cheng, Joshua Bradley, Alex Chao, Apurva Mody,
  Steven Truitt, Dasha Metropolitansky, Robert~Osazuwa Ness, and Jonathan
  Larson. 2024.
\newblock \href {https://arxiv.org/abs/2404.16130} {From local to global: A
  graph rag approach to query-focused summarization}.
\newblock \emph{arXiv preprint arXiv:2404.16130}.

\bibitem[{Gao et~al.(2023)Gao, Xiong, Gao, Jia, Pan, Bi, Dai, Sun, Wang, and
  Wang}]{Gao2023retrieval}
Yunfan Gao, Yun Xiong, Xinyu Gao, Kangxiang Jia, Jinliu Pan, Yuxi Bi, Yixin
  Dai, Jiawei Sun, Haofen Wang, and Haofen Wang. 2023.
\newblock \href {https://arxiv.org/abs/2312.10997} {Retrieval-augmented
  generation for large language models: A survey}.
\newblock \emph{arXiv preprint arXiv:2312.10997}, 2(1).

\bibitem[{Guo et~al.(2025)Guo, Yang, Zhang, Song, Zhang, Xu, Zhu, Ma, Wang, Bi
  et~al.}]{Guo2025deepseek}
Daya Guo, Dejian Yang, Haowei Zhang, Junxiao Song, Ruoyu Zhang, Runxin Xu,
  Qihao Zhu, Shirong Ma, Peiyi Wang, Xiao Bi, and 1 others. 2025.
\newblock \href {https://arxiv.org/abs/2501.12948} {Deepseek-r1: Incentivizing
  reasoning capability in llms via reinforcement learning}.
\newblock \emph{arXiv preprint arXiv:2501.12948}.

\bibitem[{Guo et~al.(2024)Guo, Xia, Yu, Ao, and Huang}]{Guo2024lightrag}
Zirui Guo, Lianghao Xia, Yanhua Yu, Tu~Ao, and Chao Huang. 2024.
\newblock \href {https://arxiv.org/abs/2410.05779} {Lightrag: Simple and fast
  retrieval-augmented generation}.
\newblock \emph{arXiv preprint arXiv:2410.05779}.

\bibitem[{Guti{\'e}rrez et~al.(2025)Guti{\'e}rrez, Shu, Qi, Zhou, and
  Su}]{Gutierrez2025from}
Bernal~Jim{\'e}nez Guti{\'e}rrez, Yiheng Shu, Weijian Qi, Sizhe Zhou, and
  Yu~Su. 2025.
\newblock \href {https://openreview.net/forum?id=LWH8yn4HS2} {From {RAG} to
  memory: Non-parametric continual learning for large language models}.
\newblock In \emph{Forty-second International Conference on Machine Learning}.

\bibitem[{Guu et~al.(2020)Guu, Lee, Tung, Pasupat, and
  Chang}]{Guu2020retrieval}
Kelvin Guu, Kenton Lee, Zora Tung, Panupong Pasupat, and Mingwei Chang. 2020.
\newblock \href {https://proceedings.mlr.press/v119/guu20a.html} {Retrieval
  augmented language model pre-training}.
\newblock In \emph{International conference on machine learning}, pages
  3929--3938. PMLR.

\bibitem[{Ho et~al.(2020)Ho, Nguyen, Sugawara, and Aizawa}]{Ho2020constructing}
Xanh Ho, Anh-Khoa~Duong Nguyen, Saku Sugawara, and Akiko Aizawa. 2020.
\newblock \href {https://aclanthology.org/2020.coling-main.580/} {Constructing
  a multi-hop qa dataset for comprehensive evaluation of reasoning steps}.
\newblock In \emph{Proceedings of the 28th International Conference on
  Computational Linguistics}, pages 6609--6625.

\bibitem[{Huang et~al.(2025)Huang, Yu, Ma, Zhong, Feng, Wang, Chen, Peng, Feng,
  Qin et~al.}]{Huang2025survey}
Lei Huang, Weijiang Yu, Weitao Ma, Weihong Zhong, Zhangyin Feng, Haotian Wang,
  Qianglong Chen, Weihua Peng, Xiaocheng Feng, Bing Qin, and 1 others. 2025.
\newblock \href {https://doi.org/10.1145/3703155} {A survey on hallucination in
  large language models: Principles, taxonomy, challenges, and open questions}.
\newblock \emph{ACM Transactions on Information Systems}, 43(2):1--55.

\bibitem[{Izacard et~al.(2022)Izacard, Caron, Hosseini, Riedel, Bojanowski,
  Joulin, and Grave}]{Izacard2022unsupervised}
Gautier Izacard, Mathilde Caron, Lucas Hosseini, Sebastian Riedel, Piotr
  Bojanowski, Armand Joulin, and Edouard Grave. 2022.
\newblock \href {https://openreview.net/forum?id=jKN1pXi7b0} {Unsupervised
  dense information retrieval with contrastive learning}.
\newblock \emph{Transactions on Machine Learning Research}.

\bibitem[{Jimenez~Gutierrez et~al.(2024)Jimenez~Gutierrez, Shu, Gu, Yasunaga,
  and Su}]{Jimenez2024hipporag}
Bernal Jimenez~Gutierrez, Yiheng Shu, Yu~Gu, Michihiro Yasunaga, and Yu~Su.
  2024.
\newblock \href
  {https://proceedings.neurips.cc/paper_files/paper/2024/file/6ddc001d07ca4f319af96a3024f6dbd1-Paper-Conference.pdf}
  {Hipporag: Neurobiologically inspired long-term memory for large language
  models}.
\newblock \emph{Advances in Neural Information Processing Systems},
  37:59532--59569.

\bibitem[{Jin et~al.(2025)Jin, Zeng, Yue, Yoon, Arik, Wang, Zamani, and
  Han}]{Jin2025search}
Bowen Jin, Hansi Zeng, Zhenrui Yue, Jinsung Yoon, Sercan Arik, Dong Wang, Hamed
  Zamani, and Jiawei Han. 2025.
\newblock \href {https://arxiv.org/abs/2503.09516} {Search-r1: Training llms to
  reason and leverage search engines with reinforcement learning}.
\newblock \emph{arXiv preprint arXiv:2503.09516}.

\bibitem[{Karpukhin et~al.(2020)Karpukhin, O{\u{g}}uz, Min, Lewis, Wu, Edunov,
  Chen, and Yih}]{Karpukhin2020dense}
Vladimir Karpukhin, Barlas O{\u{g}}uz, Sewon Min, Patrick Lewis, Ledell Wu,
  Sergey Edunov, Danqi Chen, and Wen~Tau Yih. 2020.
\newblock \href {https://aclanthology.org/2020.emnlp-main.550/} {Dense passage
  retrieval for open-domain question answering}.
\newblock In \emph{2020 Conference on Empirical Methods in Natural Language
  Processing, EMNLP 2020}, pages 6769--6781.

\bibitem[{Kwiatkowski et~al.(2019)Kwiatkowski, Palomaki, Redfield, Collins,
  Parikh, Alberti, Epstein, Polosukhin, Devlin, Lee
  et~al.}]{Kwiatkowski2019natural}
Tom Kwiatkowski, Jennimaria Palomaki, Olivia Redfield, Michael Collins, Ankur
  Parikh, Chris Alberti, Danielle Epstein, Illia Polosukhin, Jacob Devlin,
  Kenton Lee, and 1 others. 2019.
\newblock \href {https://doi.org/10.1162/tacl_a_00276} {Natural questions: A
  benchmark for question answering research}.
\newblock \emph{Transactions of the Association for Computational Linguistics},
  7:452--466.

\bibitem[{Kwon et~al.(2023)Kwon, Li, Zhuang, Sheng, Zheng, Yu, Gonzalez, Zhang,
  and Stoica}]{Kwon2023efficient}
Woosuk Kwon, Zhuohan Li, Siyuan Zhuang, Ying Sheng, Lianmin Zheng, Cody~Hao Yu,
  Joseph~E. Gonzalez, Hao Zhang, and Ion Stoica. 2023.
\newblock \href {https://doi.org/10.1145/3600006.3613165} {Efficient memory
  management for large language model serving with pagedattention}.
\newblock In \emph{Proceedings of the ACM SIGOPS 29th Symposium on Operating
  Systems Principles}.

\bibitem[{Lambert et~al.(2025)Lambert, Pyatkin, Morrison, Miranda, Lin, Chandu,
  Dziri, Kumar, Zick, Choi et~al.}]{Lambert2025rewardbench}
Nathan Lambert, Valentina Pyatkin, Jacob Morrison, Lester James~Validad
  Miranda, Bill~Yuchen Lin, Khyathi Chandu, Nouha Dziri, Sachin Kumar, Tom
  Zick, Yejin Choi, and 1 others. 2025.
\newblock \href {https://aclanthology.org/2025.findings-naacl.96/}
  {Rewardbench: Evaluating reward models for language modeling}.
\newblock In \emph{Findings of the Association for Computational Linguistics:
  NAACL 2025}, pages 1755--1797.

\bibitem[{Lee et~al.(2025{\natexlab{a}})Lee, Roy, Xu, Raiman, Shoeybi,
  Catanzaro, and Ping}]{Lee2025nvembed}
Chankyu Lee, Rajarshi Roy, Mengyao Xu, Jonathan Raiman, Mohammad Shoeybi, Bryan
  Catanzaro, and Wei Ping. 2025{\natexlab{a}}.
\newblock \href {https://openreview.net/forum?id=lgsyLSsDRe} {{NV}-embed:
  Improved techniques for training {LLM}s as generalist embedding models}.
\newblock In \emph{The Thirteenth International Conference on Learning
  Representations}.

\bibitem[{Lee et~al.(2025{\natexlab{b}})Lee, Zhu, Mavromatis, Han, Adeshina,
  Ioannidis, Rangwala, and Faloutsos}]{Lee2025hybgrag}
Meng-Chieh Lee, Qi~Zhu, Costas Mavromatis, Zhen Han, Soji Adeshina, Vassilis~N
  Ioannidis, Huzefa Rangwala, and Christos Faloutsos. 2025{\natexlab{b}}.
\newblock Hybgrag: Hybrid retrieval-augmented generation on textual and
  relational knowledge bases.
\newblock In \emph{Proceedings of the 63rd Annual Meeting of the Association
  for Computational Linguistics (Volume 1: Long Papers)}, pages 879--893.

\bibitem[{Lewis et~al.(2020)Lewis, Perez, Piktus, Petroni, Karpukhin, Goyal,
  K{\"u}ttler, Lewis, Yih, Rockt{\"a}schel et~al.}]{Lewis2020retrieval}
Patrick Lewis, Ethan Perez, Aleksandra Piktus, Fabio Petroni, Vladimir
  Karpukhin, Naman Goyal, Heinrich K{\"u}ttler, Mike Lewis, Wen-tau Yih, Tim
  Rockt{\"a}schel, and 1 others. 2020.
\newblock \href
  {https://proceedings.neurips.cc/paper_files/paper/2020/file/6b493230205f780e1bc26945df7481e5-Paper.pdf}
  {Retrieval-augmented generation for knowledge-intensive nlp tasks}.
\newblock \emph{Advances in neural information processing systems},
  33:9459--9474.

\bibitem[{Li et~al.(2025)Li, Dong, Jin, Zhang, Zhou, Zhu, Zhang, and
  Dou}]{Li2025search}
Xiaoxi Li, Guanting Dong, Jiajie Jin, Yuyao Zhang, Yujia Zhou, Yutao Zhu,
  Peitian Zhang, and Zhicheng Dou. 2025.
\newblock \href {https://arxiv.org/abs/2501.05366} {Search-o1: Agentic
  search-enhanced large reasoning models}.
\newblock \emph{arXiv preprint arXiv:2501.05366}.

\bibitem[{Mallen et~al.(2023)Mallen, Asai, Zhong, Das, Khashabi, and
  Hajishirzi}]{Mallen2023popqa}
Alex Mallen, Akari Asai, Victor Zhong, Rajarshi Das, Daniel Khashabi, and
  Hannaneh Hajishirzi. 2023.
\newblock \href {https://aclanthology.org/2023.acl-long.546/} {When not to
  trust language models: Investigating effectiveness of parametric and
  non-parametric memories}.
\newblock In \emph{Proceedings of the 61st Annual Meeting of the Association
  for Computational Linguistics (Volume 1: Long Papers)}, pages 9802--9822.

\bibitem[{Ouyang et~al.(2022)Ouyang, Wu, Jiang, Almeida, Wainwright, Mishkin,
  Zhang, Agarwal, Slama, Ray et~al.}]{Ouyang2022training}
Long Ouyang, Jeffrey Wu, Xu~Jiang, Diogo Almeida, Carroll Wainwright, Pamela
  Mishkin, Chong Zhang, Sandhini Agarwal, Katarina Slama, Alex Ray, and 1
  others. 2022.
\newblock \href
  {https://proceedings.neurips.cc/paper_files/paper/2022/file/b1efde53be364a73914f58805a001731-Paper-Conference.pdf}
  {Training language models to follow instructions with human feedback}.
\newblock \emph{Advances in neural information processing systems},
  35:27730--27744.

\bibitem[{Peng et~al.(2024)Peng, Zhu, Liu, Bo, Shi, Hong, Zhang, and
  Tang}]{Peng2024graph}
Boci Peng, Yun Zhu, Yongchao Liu, Xiaohe Bo, Haizhou Shi, Chuntao Hong, Yan
  Zhang, and Siliang Tang. 2024.
\newblock \href {https://arxiv.org/abs/2408.08921} {Graph retrieval-augmented
  generation: A survey}.
\newblock \emph{arXiv preprint arXiv:2408.08921}.

\bibitem[{Sarthi et~al.(2024)Sarthi, Abdullah, Tuli, Khanna, Goldie, and
  Manning}]{Sarthi2024raptor}
Parth Sarthi, Salman Abdullah, Aditi Tuli, Shubh Khanna, Anna Goldie, and
  Christopher~D Manning. 2024.
\newblock \href {https://openreview.net/forum?id=GN921JHCRw} {{RAPTOR}:
  Recursive abstractive processing for tree-organized retrieval}.
\newblock In \emph{The Twelfth International Conference on Learning
  Representations}.

\bibitem[{Schulman et~al.(2017)Schulman, Wolski, Dhariwal, Radford, and
  Klimov}]{Schulman2017proximal}
John Schulman, Filip Wolski, Prafulla Dhariwal, Alec Radford, and Oleg Klimov.
  2017.
\newblock \href {https://arxiv.org/abs/1707.06347} {Proximal policy
  optimization algorithms}.
\newblock \emph{arXiv preprint arXiv:1707.06347}.

\bibitem[{Shao et~al.(2024)Shao, Wang, Zhu, Xu, Song, Bi, Zhang, Zhang, Li
  et~al.}]{Shao2024deepseekmath}
Zhihong Shao, Peiyi Wang, Qihao Zhu, Runxin Xu, Junxiao Song, Xiao Bi, Haowei
  Zhang, Mingchuan Zhang, YK~Li, and 1 others. 2024.
\newblock \href {https://arxiv.org/abs/2402.03300} {Deepseekmath: Pushing the
  limits of mathematical reasoning in open language models}.
\newblock \emph{arXiv preprint arXiv:2402.03300}.

\bibitem[{Sheng et~al.(2025)Sheng, Zhang, Ye, Wu, Zhang, Zhang, Peng, Lin, and
  Wu}]{Sheng2025hybridflow}
Guangming Sheng, Chi Zhang, Zilingfeng Ye, Xibin Wu, Wang Zhang, Ru~Zhang,
  Yanghua Peng, Haibin Lin, and Chuan Wu. 2025.
\newblock \href {https://doi.org/10.1145/3689031.3696075} {Hybridflow: A
  flexible and efficient rlhf framework}.
\newblock In \emph{Proceedings of the Twentieth European Conference on Computer
  Systems}, pages 1279--1297.

\bibitem[{Song et~al.(2025)Song, Jiang, Min, Chen, Chen, Zhao, Fang, and
  Wen}]{Song2025r1}
Huatong Song, Jinhao Jiang, Yingqian Min, Jie Chen, Zhipeng Chen, Wayne~Xin
  Zhao, Lei Fang, and Ji-Rong Wen. 2025.
\newblock \href {https://arxiv.org/abs/2503.05592} {R1-searcher: Incentivizing
  the search capability in llms via reinforcement learning}.
\newblock \emph{arXiv preprint arXiv:2503.05592}.

\bibitem[{Team et~al.(2024)Team, Georgiev, Lei, Burnell, Bai, Gulati, Tanzer,
  Vincent, Pan, Wang et~al.}]{Team2024gemini}
Gemini Team, Petko Georgiev, Ving~Ian Lei, Ryan Burnell, Libin Bai, Anmol
  Gulati, Garrett Tanzer, Damien Vincent, Zhufeng Pan, Shibo Wang, and 1
  others. 2024.
\newblock \href {https://arxiv.org/abs/2403.05530} {Gemini 1.5: Unlocking
  multimodal understanding across millions of tokens of context}.
\newblock \emph{arXiv preprint arXiv:2403.05530}.

\bibitem[{Touvron et~al.(2023)Touvron, Martin, Stone, Albert, Almahairi,
  Babaei, Bashlykov, Batra, Bhargava, Bhosale et~al.}]{Touvron2023llama}
Hugo Touvron, Louis Martin, Kevin Stone, Peter Albert, Amjad Almahairi, Yasmine
  Babaei, Nikolay Bashlykov, Soumya Batra, Prajjwal Bhargava, Shruti Bhosale,
  and 1 others. 2023.
\newblock \href {https://arxiv.org/abs/2307.09288} {Llama 2: Open foundation
  and fine-tuned chat models}.
\newblock \emph{arXiv preprint arXiv:2307.09288}.

\bibitem[{Trivedi et~al.(2022)Trivedi, Balasubramanian, Khot, and
  Sabharwal}]{Trivedi2022musique}
Harsh Trivedi, Niranjan Balasubramanian, Tushar Khot, and Ashish Sabharwal.
  2022.
\newblock \href {https://doi.org/10.1162/tacl_a_00475} {Musique: Multihop
  questions via single-hop question composition}.
\newblock \emph{Transactions of the Association for Computational Linguistics},
  10:539--554.

\bibitem[{Trivedi et~al.(2023)Trivedi, Balasubramanian, Khot, and
  Sabharwal}]{Trivedi2023interleaving}
Harsh Trivedi, Niranjan Balasubramanian, Tushar Khot, and Ashish Sabharwal.
  2023.
\newblock \href {https://aclanthology.org/2023.acl-long.557/} {Interleaving
  retrieval with chain-of-thought reasoning for knowledge-intensive multi-step
  questions}.
\newblock In \emph{The 61st Annual Meeting Of The Association For Computational
  Linguistics}.

\bibitem[{Wang et~al.(2023)Wang, Li, Sun, and Liu}]{Wang2023self}
Yile Wang, Peng Li, Maosong Sun, and Yang Liu. 2023.
\newblock \href {https://aclanthology.org/2023.findings-emnlp.691/}
  {Self-knowledge guided retrieval augmentation for large language models}.
\newblock In \emph{Findings of the Association for Computational Linguistics:
  EMNLP 2023}, pages 10303--10315.

\bibitem[{Wang et~al.(2024)Wang, Ren, Li, Zhao, Liu, and Wen}]{Wang2024rear}
Yuhao Wang, Ruiyang Ren, Junyi Li, Wayne~Xin Zhao, Jing Liu, and Ji-Rong Wen.
  2024.
\newblock \href {https://aclanthology.org/2024.emnlp-main.321/} {Rear: A
  relevance-aware retrieval-augmented framework for open-domain question
  answering}.
\newblock In \emph{Proceedings of the 2024 Conference on Empirical Methods in
  Natural Language Processing}, pages 5613--5626.

\bibitem[{Wei et~al.(2022)Wei, Wang, Schuurmans, Bosma, Xia, Chi, Le, Zhou
  et~al.}]{Wei2022chain}
Jason Wei, Xuezhi Wang, Dale Schuurmans, Maarten Bosma, Fei Xia, Ed~Chi, Quoc~V
  Le, Denny Zhou, and 1 others. 2022.
\newblock \href
  {https://proceedings.neurips.cc/paper_files/paper/2022/file/9d5609613524ecf4f15af0f7b31abca4-Paper-Conference.pdf}
  {Chain-of-thought prompting elicits reasoning in large language models}.
\newblock \emph{Advances in neural information processing systems},
  35:24824--24837.

\bibitem[{Yang et~al.(2025{\natexlab{a}})Yang, Li, Yang, Zhang, Hui, Zheng, Yu,
  Gao, Huang, Lv et~al.}]{Yang2025qwen3}
An~Yang, Anfeng Li, Baosong Yang, Beichen Zhang, Binyuan Hui, Bo~Zheng, Bowen
  Yu, Chang Gao, Chengen Huang, Chenxu Lv, and 1 others. 2025{\natexlab{a}}.
\newblock \href {https://arxiv.org/abs/2505.09388} {Qwen3 technical report}.
\newblock \emph{arXiv preprint arXiv:2505.09388}.

\bibitem[{Yang et~al.(2025{\natexlab{b}})Yang, Yang, Zhang, Hui, Zheng, Yu, Li,
  Liu, Huang, Wei, Lin, Yang, Tu, Zhang, Yang, Yang, Zhou, Lin, Dang, Lu, Bao,
  Yang, Yu, Li, Xue, Zhang, Zhu, Men, Lin, Li, Tang, Xia, Ren, Ren, Fan, Su,
  Zhang, Wan, Liu, Cui, Zhang, and Qiu}]{Qwen2025qwen25}
An~Yang, Baosong Yang, Beichen Zhang, Binyuan Hui, Bo~Zheng, Bowen Yu,
  Chengyuan Li, Dayiheng Liu, Fei Huang, Haoran Wei, Huan Lin, Jian Yang,
  Jianhong Tu, Jianwei Zhang, Jianxin Yang, Jiaxi Yang, Jingren Zhou, Junyang
  Lin, Kai Dang, and 23 others. 2025{\natexlab{b}}.
\newblock \href {https://arxiv.org/abs/2412.15115} {Qwen2.5 technical report}.
\newblock \emph{arXiv preprint arXiv:2412.15115}.

\bibitem[{Yang et~al.(2018)Yang, Qi, Zhang, Bengio, Cohen, Salakhutdinov, and
  Manning}]{Yang2018hotpotqa}
Zhilin Yang, Peng Qi, Saizheng Zhang, Yoshua Bengio, William Cohen, Ruslan
  Salakhutdinov, and Christopher~D Manning. 2018.
\newblock \href {https://aclanthology.org/D18-1259/} {Hotpotqa: A dataset for
  diverse, explainable multi-hop question answering}.
\newblock In \emph{Proceedings of the 2018 Conference on Empirical Methods in
  Natural Language Processing}, pages 2369--2380.

\bibitem[{Yao et~al.(2023)Yao, Zhao, Yu, Du, Shafran, Narasimhan, and
  Cao}]{Yao2023react}
Shunyu Yao, Jeffrey Zhao, Dian Yu, Nan Du, Izhak Shafran, Karthik Narasimhan,
  and Yuan Cao. 2023.
\newblock \href {https://openreview.net/forum?id=WE_vluYUL-X} {React:
  Synergizing reasoning and acting in language models}.
\newblock In \emph{The Eleventh International Conference on Learning
  Representations}.

\bibitem[{Yu et~al.(2022)Yu, Zhu, Li, Hu, Wang, Ji, and Jiang}]{Yu2022survey}
Wenhao Yu, Chenguang Zhu, Zaitang Li, Zhiting Hu, Qingyun Wang, Heng Ji, and
  Meng Jiang. 2022.
\newblock \href {https://doi.org/10.1145/3512467} {A survey of
  knowledge-enhanced text generation}.
\newblock \emph{ACM Computing Surveys}, 54(11s):1--38.

\bibitem[{Zhao et~al.(2023)Zhao, Zhou, Li, Tang, Wang, Hou, Min, Zhang, Zhang,
  Dong et~al.}]{Zhao2023survey}
Wayne~Xin Zhao, Kun Zhou, Junyi Li, Tianyi Tang, Xiaolei Wang, Yupeng Hou,
  Yingqian Min, Beichen Zhang, Junjie Zhang, Zican Dong, and 1 others. 2023.
\newblock \href {https://arxiv.org/abs/2303.18223} {A survey of large language
  models}.
\newblock \emph{arXiv preprint arXiv:2303.18223}, 1(2).

\bibitem[{Zhou et~al.(2023)Zhou, Lu, Mishra, Brahma, Basu, Luan, Zhou, and
  Hou}]{Zhou2023instruction}
Jeffrey Zhou, Tianjian Lu, Swaroop Mishra, Siddhartha Brahma, Sujoy Basu,
  Yi~Luan, Denny Zhou, and Le~Hou. 2023.
\newblock \href {https://arxiv.org/abs/2311.07911} {Instruction-following
  evaluation for large language models}.
\newblock \emph{arXiv preprint arXiv:2311.07911}.

\end{thebibliography}

\appendix
\section{Analysis on Efficiency Reward}
\label{appendix:efficiency_reward_analysis}
In this section, we provide a detailed theoretical analysis of why the efficiency reward designed in \model{} improves selective retrieval in GRPO-based training.

\subsection{Batch-Level Efficiency Reward}
Let $\tau_i$ denote the $i$-th trajectory in a group of $G$ trajectories sampled from a batch of size $B$. The total reward for trajectory $\tau_i$ is
\begin{equation}
\small
R_{\phi}(\tau_i) =
\begin{cases}
R_{\text{outcome}}(\tau_i),\!\!&\!\!R_{\text{outcome}}(\tau_i)=0\\
R_{\text{outcome}}(\tau_i) + R_{\text{efficiency}}(\tau_i),\!\!&\!\!R_{\text{outcome}}(\tau_i)=1
\end{cases},
\end{equation}
where $R_{\text{outcome}}(\tau_i) \in \{0,1\}$ indicates the correctness of the answer.
The efficiency reward is centered on the batch-level average retrieval time rather than the group-level average:
\begin{equation}
R_{\text{efficiency}}(\tau_i) = \frac{t_{\text{avg}} - t_i}{T}, \quad 
t_{\text{avg}} = \frac{1}{B} \sum_{\tau \in \text{batch}} t_\tau,
\end{equation}
where $t_i$ is the total retrieval time of trajectory $\tau_i$, and $T$ is a normalization constant.

There are three main reasons for choosing batch-level efficiency reward together with GRPO-based training, instead of using group-level efficiency:
\begin{itemize}
    \item \textbf{Variance reduction and stability.} Batch-level averaging reduces the impact of noisy fluctuations in retrieval time (e.g., hardware latency or network delays).  
    \item \textbf{Mitigating anomalies across queries.} Although batch-level normalization may produce unusually high or low raw efficiency rewards for certain queries, GRPO’s group-relative advantage compensates for this effect.  
    \item \textbf{Encouraging selective retrieval.} Combining batch-level centering with GRPO advantage ensures that trajectories with unnecessary retrieval are penalized, while efficient yet accurate trajectories are favored.
\end{itemize}

\subsection{Variance Reduction and Stability}
Each GRPO group may contain only a few trajectories (e.g., $G=5$). Raw retrieval times $t_i$ can fluctuate due to hardware noise, network latency, or retriever stochasticity.  
If group-level averaging were used, such fluctuations could lead to unstable rewards.  
By computing $t_{\text{avg}}$ across the entire batch, we obtain a more stable reference signal that smooths out these random variations.  

This reduces the variance of the group-relative advantage, which in turn stabilizes policy gradient updates.

\subsection{Mitigating Anomalies Across Queries}
Batch-level normalization may produce unusually high or low raw efficiency rewards for certain queries.  
For example, a simple question requiring little retrieval could yield a disproportionately large positive $R_{\text{efficiency}}(\tau_i)$.  
This is indeed a potential drawback of using batch-level efficiency reward.

However, GRPO compensates for this issue through its group-relative formulation.  
Although $R_{\text{efficiency}}(\tau_i)$ is normalized at the batch level, the advantage $A_i$ is computed relative to the group mean reward within each GRPO group, i.e.,
\begin{equation}
\label{eq:Ai}
A_i \;=\;
\frac{R_\phi(\tau_i) - \frac{1}{G}\sum_{j=1}^G R_\phi(\tau_j)}
{\mathrm{std}\big(\{R_\phi(\tau_j)\}_{j=1}^G\big)}.
\end{equation}

Even if a particular query obtains an abnormally large batch-normalized efficiency reward, its influence on learning is moderated by this group-relative centering.  
As a result, the combination of batch-level normalization and group-level centering ensures that the learning signal remains consistent across diverse query types.

\subsection{Encouraging Selective Retrieval}
\label{appendix:selective_retrieval}
For trajectories that correctly answer the query ($R_{\text{outcome}}=1$), the numerator of $A_i$ can be decomposed into outcome and efficiency components, yielding
\begin{equation}
\label{eq:Ai_decomp}
\small
A_i \;=\; 
\frac{\big(1 - \overline{R_{\text{outcome}}}\big) + \big(R_{\text{efficiency}}(\tau_i) - \overline{R_{\text{efficiency}}}\big)}
{\mathrm{std}\big(\{R_\phi(\tau_j)\}_{j=1}^G\big)},
\end{equation}
where $\overline{R_{\text{outcome}}}$ and $\overline{R_{\text{efficiency}}}$ are group means computed within the GRPO group, while $R_{\text{efficiency}}(\tau_i)$ itself is computed using the batch-level reference $t_{\mathrm{avg}}$.  

From \Cref{eq:Ai_decomp} we see:
\begin{itemize}
  \item If a trajectory answers correctly and its retrieval time is less than the group average, then $R_{\text{efficiency}}(\tau_i)-\overline{R_{\text{efficiency}}}>0$, and consequently, the numerator increases, thereby rewarding the policy for selective retrieval.
  \item If a trajectory performs redundant retrieval, the efficiency term is negative and reduces $A_i$, discouraging unnecessary retrieval.
\end{itemize}

Thus, GRPO guides the policy towards trajectories that balance correctness with efficient retrieval.

\section{Implementation Details}
\label{appendix:implementation_details}
\subsection{Evaluation Datasets and Baselines}
\textbf{Evaluation Datasets.} The statistics of evaluation datasets are shown in~\Cref{tab:dataset_statistics}.
\begin{table}[ht]
\centering
\caption{Dataset statistics}
\label{tab:dataset_statistics}
\begin{tabular}{lcc}
\toprule
\textbf{Dataset} & \textbf{\# of queries} & \textbf{\# of passages} \\
\midrule
PopQA & 1,000 & 8,678 \\
NQ & 1,000 & 9,633 \\
HotpotQA & 1,000 & 9,811 \\
2Wiki & 1,000 & 6,119 \\
MuSiQue & 1,000 & 11,656 \\
\bottomrule
\end{tabular}
\end{table}

\noindent \textbf{Baselines.}
\begin{itemize}
    \item \textbf{Vanilla RAG}~\citep{Lewis2020retrieval} is a standard RAG framework that combines an LLM with a learned retriever to condition generation on retrieved documents.
    \item \textbf{Search-o1}~\citep{Li2025search} is an agentic retrieval system that performs on-demand multi-step search with a ``reason-in-documents'' refinement module.
    \item \textbf{Search-R1}~\citep{Jin2025search} is an RL-trained model that interleaves stepwise reasoning with autonomous search actions.
    \item \textbf{R1-Searcher}~\citep{Song2025r1} is an RL approach that teaches LLMs when and how to invoke external search for improved reasoning.
    \item \textbf{GraphRAG}~\citep{Edge2024local} is a graph-based RAG method that aggregates evidence via local-to-global graph traversal.
    \item \textbf{LightRAG}~\citep{Guo2024lightrag} is a lightweight and efficient graph-based RAG method with simplified indexing and fast dual-level retrieval.
    \item \textbf{RAPTOR}~\citep{Sarthi2024raptor} is a hierarchical, tree-structured RAG approach that uses recursive summarization to access multi-level abstractions.
    \item \textbf{HippoRAG}~\citep{Jimenez2024hipporag} is a memory-inspired RAG framework that builds a knowledge graph and retrieves via personalized PageRank.
    \item \textbf{HippoRAG 2}~\citep{Gutierrez2025from} is an enhanced graph-based RAG framework that extends HippoRAG with deeper passage integration.
\end{itemize}

\subsection{Training Prompt Template}
The training prompt template for the policy LLM is shown in Table~\ref{tab:prompt_template}.
\begin{table*}[h]
\caption{Training prompt template for multi-turn reasoning and retrieval.}
\label{tab:prompt_template}
\centering
\begin{tcolorbox}[
    colback=black!5, 
    colframe=black!70!white, 
    title=\textbf{Training Prompt Template for the Policy LLM},
    fonttitle=\small\bfseries,
    arc=3mm, 
    boxrule=0.5pt 
]
\small
\begin{tabularx}{\textwidth}{X}
Answer the given question. You must conduct reasoning inside \think{think} and \think{/think} first every time you get new information. After reasoning, if you find you lack some knowledge, you can call a search engine using the following strict format: \\
1. You MUST first decide which retrieval mode to use (both modes will return relevant documents, but use different retrieval methods):
\begin{itemize}[leftmargin=16pt]
    \item[-] Use \passage{} to find documents using semantic similarity-based dense retrieval
    \item[-] Use \graph{} to find documents through graph-based retrieval, which performs retrieval on a structured knowledge graph constructed from documents using fact ranking and graph reasoning
    \item[-] You can combine them as \graph{}\passage{} to get documents from both retrieval methods
\end{itemize}
2. Then formulate your specific search query based on what information you need \\
3. Finally, wrap everything in \search{search} and \search{/search} tags \\
For example:
\begin{itemize}[leftmargin=16pt]
    \item[-] Using dense retrieval: \search{search} \passage{} the capital of France \search{/search}
    \item[-] Using graph-based retrieval: \search{search} \graph{} the capital of France \search{/search}
    \item[-] Using both methods: \search{search} \graph{}\passage{} the capital of France \search{/search}
\end{itemize}
The search results (relevant documents) will be returned between \info{information} and \info{/information} tags. You can search as many times as you want. If you find no further external knowledge needed, you can directly provide the answer inside \answer{answer} and \answer{/answer}, without detailed illustrations. For example, \answer{answer} Paris \answer{/answer}. \\
Question: \{question\}
\end{tabularx}
\end{tcolorbox}
\end{table*}

\subsection{Training Details}
\textbf{Hyperparameters.}  
For GRPO training of \model{}, we set the policy LLM learning rate to $1\times 10^{-6}$, a total batch size of 256, with a mini-batch size of 128 and a micro-batch size of 32. The KL divergence regularization coefficient $\beta$ is set to 0.001, and the clip ratio $\epsilon$ is set to 0.2. The retrieval budget is fixed at $B=4$, and the number of retrieved passages per call is $k=3$. The maximum sequence length is set to 4,096 tokens, with a maximum response length of 500 tokens, a maximum start length of 2,048 tokens, and a maximum observation length of 500 tokens.

\noindent \textbf{Open Information Extraction Models}
We employ Llama-3.1-8B-Instruct and Llama-3.3-70B-Instruct for Open Information Extraction (OpenIE) during graph construction. Following HippoRAG 2, the model is used to extract entities and relational triplets from corpus passages, which are then used to build the knowledge graph that supports graph and hybrid retrieval in both HippoRAG 2 and \model{}.
For experiments based on 3B models, we use Llama-3.1-8B-Instruct as a more lightweight alternative to the larger OpenIE models (Llama-3.3-70B-Instruct and GPT-4o-mini) adopted in HippoRAG 2. 
For experiments involving 7B models, we adopt the same Llama-3.3-70B-Instruct model used in HippoRAG 2 to construct higher-quality graphs.

\noindent \textbf{Training Configuration.}  
Our training framework is adapted from the Search-R1 training framework~\citep{Jin2025search}, which builds upon the verl~\citep{Sheng2025hybridflow}. Training is conducted on a single node with 8$\times$80GB NVIDIA A100 GPUs. To improve memory efficiency, we enable gradient checkpointing and apply Fully Sharded Data Parallel (FSDP) with CPU offloading for parameters, gradients, and optimizer states. Rollouts are sampled with vLLM~\citep{Kwon2023efficient} using a tensor parallel size of 1 and a GPU memory utilization ratio of 0.6. The rollout sampling temperature is set to 1.0.

\noindent \textbf{Two-Stage Training.}  
Stage~1 is trained for 20 steps (0.5 epoch) with EM-based rewards only, ensuring correctness. Stage~2 continues for an additional 20 steps (0.5 epoch) with the accuracy--efficiency reward introduced in~\Cref{sec:two_stage_RL}. In both stages, we sample five responses per prompt during training to compute group-relative advantages. Checkpoints are saved every 10 steps, and the final checkpoint is used for evaluation.

\begin{table*}[t]
\centering
\small
\renewcommand{\arraystretch}{1.2}
\resizebox{\textwidth}{!}{%
\begin{tabular}{lcccccccccccc}
\toprule
\multirow{3}{*}{\textbf{Method}} 
& \multicolumn{4}{c}{\textbf{Simple QA}} 
& \multicolumn{6}{c}{\textbf{Multi-hop QA}} 
& \multicolumn{2}{c}{\multirow{2}{*}{\textbf{Average}}}\\
\cmidrule(lr){2-5} \cmidrule(lr){6-11}
 & \multicolumn{2}{c}{\textbf{PopQA}} 
 & \multicolumn{2}{c}{\textbf{NQ}} 
 & \multicolumn{2}{c}{\textbf{HotpotQA}} 
 & \multicolumn{2}{c}{\textbf{2Wiki}} 
 & \multicolumn{2}{c}{\textbf{MuSiQue}}
 & & \\
\cmidrule(lr){2-3} \cmidrule(lr){4-5} \cmidrule(lr){6-7} \cmidrule(lr){8-9} \cmidrule(lr){10-11} \cmidrule(lr){12-13}
 & \textbf{EM} & \textbf{F1} 
 & \textbf{EM} & \textbf{F1} 
 & \textbf{EM} & \textbf{F1} 
 & \textbf{EM} & \textbf{F1} 
 & \textbf{EM} & \textbf{F1}
 & \textbf{EM} & \textbf{F1}\\
\midrule
\rowcolor{gray!10}
\multicolumn{13}{c}{{\centering\raisebox{-.3\height}{\includegraphics[width=0.4cm]{figs/logo/qwen.png}}} \textit{\textbf{Qwen2.5-3B}}}\\
\midrule
\multicolumn{13}{l}{\textit{\textbf{Contriever}}}\\
Search-R1 & 45.8 & 53.3 & 46.2$^*$ & 54.8$^*$ & 45.2$^*$ & 56.9$^*$ & 42.4 & 50.8 & 22.2 & 30.9 & 40.4 & 49.3 \\
\model{} & 49.4 & 56.8 & 44.1 & 53.4 & \underline{53.2}$^*$ & \underline{65.1}$^*$ & \underline{57.5} & \underline{64.1} & 30.7 & 39.3 & 47.0 & 55.7 \\
\multicolumn{13}{l}{\textit{\textbf{NV-Embed-v2}}}\\
Search-R1 & \textbf{53.5} & \textbf{60.1} & \textbf{54.8}$^*$ & \textbf{63.1}$^*$ & 51.0$^*$ & 62.5$^*$ & 45.7 & 54.5 & \underline{32.1} & \underline{41.0} & \underline{47.4} & \underline{56.2} \\
\model{} & \underline{53.0} & \underline{59.6} & \underline{47.0} & \underline{57.3} & \textbf{56.1}$^*$ & \textbf{68.2}$^*$ & \textbf{58.2} & \textbf{65.4} & \textbf{36.6} & \textbf{45.4} & \textbf{50.2} & \textbf{59.2} \\
\midrule
\rowcolor{gray!10}
\multicolumn{13}{c}{{\centering\raisebox{-.3\height}{\includegraphics[width=0.4cm]{figs/logo/qwen.png}}} \textit{\textbf{Qwen2.5-7B}}}\\
\midrule
\multicolumn{13}{l}{\textit{\textbf{Contriever}}}\\
Search-R1 & 47.1 & 52.7 & 49.5$^*$ & 58.1$^*$ & 45.2$^*$ & 55.7$^*$ & 45.9 & 52.7 & 24.4 & 31.8 & 42.4 & 50.2 \\
\model{} & 50.5 & \underline{56.7} & 46.9 & 55.4 & \underline{53.8}$^*$ & \underline{65.3}$^*$ & \underline{55.3} & \underline{62.7} & \underline{34.3} & \underline{43.7} & 48.2 & \underline{56.8} \\
\multicolumn{13}{l}{\textit{\textbf{NV-Embed-v2}}}\\
Search-R1 & \textbf{51.3} & \textbf{57.1} & \textbf{56.8}$^*$ & \textbf{65.3}$^*$ & 51.0$^*$ & 62.0$^*$ & 51.8 & 58.9 & 32.0 & 40.8 & \underline{48.6} & \underline{56.8} \\
\model{} & \underline{50.6} & 56.4 & \underline{51.5} & \underline{60.4} & \textbf{60.8}$^*$ & \textbf{72.5}$^*$ & \textbf{57.1} & \textbf{64.6} & \textbf{39.6} & \textbf{49.3} & \textbf{51.9} & \textbf{60.6} \\
\bottomrule
\end{tabular}}
\caption{Performance comparison among \model{} and Search-R1 with different dense retrievers. $^*$ represents in-domain datasets.}
\label{tab:retriever}
\end{table*}
\begin{figure*}
    \centering
    \includegraphics[width=\linewidth]{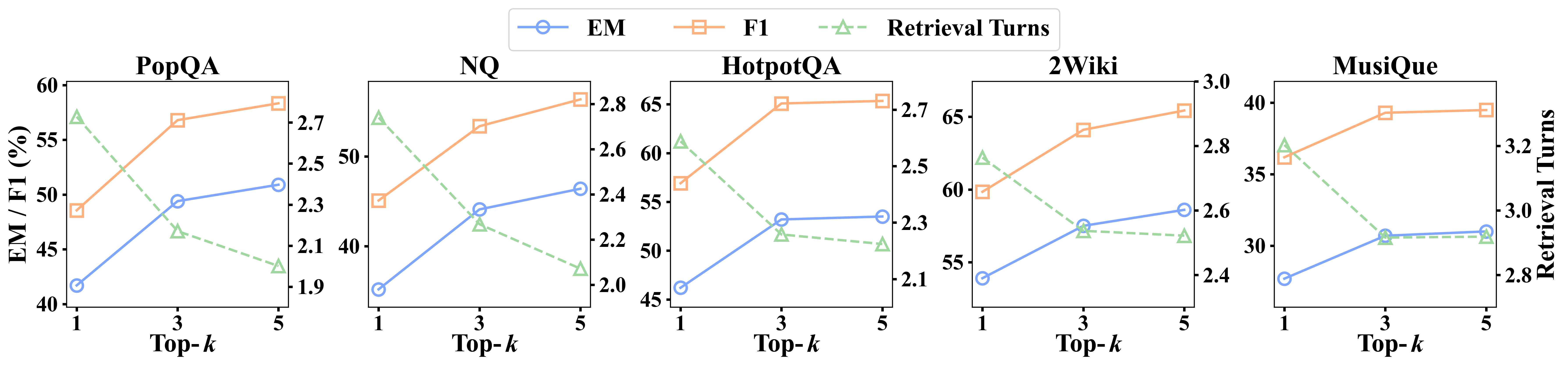}
    \caption{Performance of \model{}-3B with different number of retrieved documents.}
    \label{fig:topk}
\end{figure*}
\section{Additional Experiments}
\subsection{Analysis on Dense Retrievers}
\Cref{tab:retriever} shows the performance of \model{} and Search-R1 with two different dense retrievers, i.e., Contriever~\citep{Izacard2022unsupervised} and NV-Embed-v2~\citep{Lee2025nvembed}.
Across both 3B and 7B model sizes, \model{} yields consistent gains on multi-hop QA benchmarks, regardless of which dense retriever is used. In contrast, Search-R1 exhibits more sensitive to the quality of the dense retriever, benefiting substantially from the stronger NV-Embed-v2 retriever on simple QA tasks.
This highlights that \model{} relies less heavily on the dense retriever, since its adaptive use of graph and hybrid retrieval provides acurate evidence, thereby stabilizing performance even when the dense retriever is relatively weak.

\subsection{Analysis on Number of Retrieved Documents}
In this experiment, we analyze the performance of \model{}-3B with varying numbers of retrieved documents. As shown in \Cref{fig:topk}, the increasing $k$ generally improves both EM and F1 across all datasets, though the gains diminish beyond $k=3$. Interestingly, retrieval turns consistently decrease as $k$ grows, indicating that providing the model with richer evidence reduces the need for iterative retrieval. These results demonstrate that a moderate retrieval breadth yields a balance between accuracy and retrieval efficiency.

\section{Case Study}
\label{appendix:case_study}
To further illustrate how our training framework improves model behavior, we present several qualitative case studies comparing the outputs of the base model before training and our proposed \model{} after training, as shown in Tables~\ref{tab:case_study_before_after_1}--\ref{tab:case_study_before_after_2}. 
These cases reveal several representative issues of the base model and how our training addresses them.

\textbf{(1) Over-reliance on internal knowledge without validation}. The base model tended to rely on its parametric knowledge and give confident but unsupported answers, skipping the step of verifying correctness against external evidence (Case 1). Our model, by contrast, learns to cross-check candidate answers with retrieved documents before finalizing its response, which reduces hallucinations and increases factual accuracy.

\textbf{(2) Failure to understand or decompose complex questions}. The base model frequently struggled with multi-hop or compositional questions, failing to recognize their underlying structure or to plan effective retrieval steps (Case 2). Through training, our model acquires the ability to analyze the question, break it down into subproblems, and issue precise retrieval queries for each part, which leads to more accurate and complete answers.

Overall, after training, \model{} shows clear improvements in its ability to understand complex questions, decompose them into manageable subproblems, and construct concise yet targeted retrieval queries that lead to well-grounded and correct answers.

\begin{table*}[htbp]
\caption{\textbf{Case study 1}. Before training, the model hallucinates incorrect internal knowledge, thus wrongly claiming that Johnny Pemberton played Bo Thompson in ``That '70s Show'' and that it was created by Steven Molaro. After training, it avoids hallucinations, correctly decomposes the question into sub-steps, and retrieves the right evidence to identify the creator as Justin Spitzer.}
\label{tab:case_study_before_after_1}
\centering
\small
\renewcommand{\arraystretch}{1.2}
\begin{tabularx}{\textwidth}{lX}
\toprule
\textbf{Question}: Who created the NBC sitcom that Johnny Pemberton appears in as the character Bo Thompson? \\
\midrule
\textbf{Ground Truth}: Justin Spitzer \\
\midrule

\multicolumn{2}{l}{\textbf{Before Training (Qwen2.5-3B-Instruct)}} \\

\multicolumn{2}{X}{%
\parbox{\hsize}{%
\think{think} To answer this question, I need to identify the NBC sitcom that Johnny Pemberton appeared in as the character Bo Thompson. Let's start with the reasoning: 1. Johnny Pemberton is known for his role as Bo Thompson in the NBC sitcom ``That '70s Show.'' 2. ``That '70s Show'' aired from 1998 to 2006. 3. The show was created by Steven Molaro, who wrote the pilot and served as showrunner. Now, I will search for more detailed information about the creator of ``That '70s Show'' to confirm the answer. \think{/think}\\
\search{search} \passage{} Steven Molaro That '70s Show creator \search{/search}\\
\info{information} Doc 1(Title: Bigfoot (TV series)) Bigfoot is an American animated series television pilot created by Seth Rogen, Matt McKenna, and Evan Goldberg... \info{/information}
\par\noindent\hdashrule[0.5ex]{\linewidth}{1pt}{2pt}\par
\think{think} Based on the search results, the information about Steven Molaro and ``That '70s Show'' is not found in the documents. However, the documents do not contain any information about Johnny Pemberton or the show ``That '70s Show.'' Since the documents do not provide the necessary information, I will use graph-based retrieval to find the creator of ``That '70s Show.'' \think{/think}\\
\search{search} \graph{} That '70s Show creator \search{/search}\\
\info{information} Doc 1(Title: Parent show) Refers to a TV series that has spawned a spin-off... \info{/information}
\par\noindent\hdashrule[0.5ex]{\linewidth}{1pt}{2pt}\par
\think{think} The search results do not provide the creator of ``That '70s Show.'' Given the lack of relevant information, I will conclude based on the initial reasoning: \think{/think}\\
\answer{answer} Steven Molaro \answer{/answer}
}}\\
\midrule
\multicolumn{2}{l}{\textbf{After Training (\model{}-3B)}} \\
\multicolumn{2}{X}{%
\parbox{\hsize}{%
\think{think} To answer this question, I need to first identify the NBC sitcom Johnny Pemberton appears in as the character Bo Thompson. Then, I'll find out who created that show. Let's start with the sitcom. \think{/think}\\
\search{search} \graph{} Johnny Pemberton Bo Thompson \search{/search}\\
\info{information} Doc 1(Title: Johnny Pemberton) Johnny Pemberton (born in 1981) is an American actor and comedian from Rochester, Minnesota. He is best known for his role as the titular ``Son of Zorn'' in the short-lived Fox sitcom, and has also appeared as the recurring character Bo Thompson in the NBC sitcom ``Superstore''... \info{/information}
\par\noindent\hdashrule[0.5ex]{\linewidth}{1pt}{2pt}\par
\think{think} From Doc 1, it's clear that Johnny Pemberton appeared as Bo Thompson in ``Superstore''. Now, I need to find out who created ``Superstore''. \think{/think}\\
\search{search} \graph{} Superstore creator \search{/search}\\
\info{information} Doc 1(Title: Superstore (TV series)) Superstore is an American single-camera sitcom television series that premiered on NBC on November 30, 2015. The series was created by Justin Spitzer... \info{/information}
\par\noindent\hdashrule[0.5ex]{\linewidth}{1pt}{2pt}\par
\think{think} From Doc 1, it's evident that ``Superstore'' was created by Justin Spitzer. Therefore, the answer to the question is: \think{/think}\\
\answer{answer} Justin Spitzer \answer{/answer}
}}\\
\bottomrule
\end{tabularx}
\end{table*}

\begin{table*}[ht]
\caption{\textbf{Case study 2}. Before training, the model failed to answer the question due to a lack of reasoning and question decomposition. After training, the model correctly decomposed the question into two steps, i.e., identifying the performer of ``Hello Tomorrow'' as Dave Koz, and then finding out who was featured on Smooth Jazz Stations along with Kenny G and Dave Koz.}
\label{tab:case_study_before_after_2}
\centering
\small
\renewcommand{\arraystretch}{1.2}
\begin{tabularx}{\textwidth}{lX}
\toprule
\textbf{Question}: Along with Kenny G and the performer of Hello Tomorrow, what artist was featured on Smooth Jazz Stations? \\
\midrule
\textbf{Ground Truth}: George Benson \\
\midrule

\multicolumn{2}{l}{\textbf{Before Training (Qwen2.5-7B-Instruct)}} \\

\multicolumn{2}{X}{%
\parbox{\hsize}{%
\think{think} I need to find out which artist was featured on Smooth Jazz Stations along with Kenny G and the performer of Hello Tomorrow. I don't have this information directly, so I will need to search for it. \think{/think}\\
\search{search} \graph{}\passage{} artists featured on Smooth Jazz Stations with Kenny G and Hello Tomorrow performer \search{/search}\\
\info{information} Doc 1(Title: Adult contemporary music) In its early years of existence, the smooth jazz format...artists like George Benson, Kenny G and Dave Koz...\\
Doc 2(Title: Hello Tomorrow (album)) Hello Tomorrow is the ninth studio album by saxophone player Dave Koz... \info{/information}
\par\noindent\hdashrule[0.5ex]{\linewidth}{1pt}{2pt}\par
\answer{answer} Dave Koz \answer{/answer}
}}\\
\midrule
\multicolumn{2}{l}{\textbf{After Training (\model{}-7B)}} \\
\multicolumn{2}{X}{%
\parbox{\hsize}{%
\think{think} To answer this question, I need to identify three artists: Kenny G, the performer of ``Hello Tomorrow'', and the artist featured on Smooth Jazz Stations. I will start by identifying the performer of ``Hello Tomorrow''. \think{/think}\\
\search{search} \graph{} performer of Hello Tomorrow \search{/search}\\
\info{information} Doc 1(Title: Hello Tomorrow (album)) Hello Tomorrow is the ninth studio album by saxophone player Dave Koz...\info{/information}
\par\noindent\hdashrule[0.5ex]{\linewidth}{1pt}{2pt}\par
\think{think} From the information provided, Dave Koz performed ``Hello Tomorrow''. Now I need to find out who was featured on Smooth Jazz Stations along with Kenny G and Dave Koz. \think{/think}\\
\search{search} \graph{}\passage{} artist featured on Smooth Jazz Stations along with Kenny G and Dave Koz \search{/search}\\
\info{information} Doc 1(Title: Adult contemporary music) In its early years of existence, the smooth jazz format...artists like George Benson, Kenny G and Dave Koz...\info{/information}
\par\noindent\hdashrule[0.5ex]{\linewidth}{1pt}{2pt}\par
\think{think} From the information, it seems that the artist featured on Smooth Jazz Stations along with Kenny G and Dave Koz is George Benson. \think{/think}\\
\answer{answer} George Benson \answer{/answer}
}}\\
\bottomrule
\end{tabularx}
\end{table*}

\end{document}